\documentclass{article}

\usepackage{PRIMEarxiv}

\usepackage[utf8]{inputenc} 
\usepackage[T1]{fontenc}    
\usepackage{hyperref}       
\usepackage{url}            
\usepackage{booktabs}       
\usepackage{amsfonts}       
\usepackage{nicefrac}       
\usepackage{microtype}      
\usepackage{lipsum}
\usepackage{graphicx}
\graphicspath{{media/}}     
\newcommand{\vect}{\bm}
\newcommand{\mat}{\bm}
\newcommand{\ope}{\mathbb}
\newcommand{\vectt}{\mathbf}

\newcommand{\field}{\mathbf}
\usepackage[table]{xcolor}

\usepackage{graphicx}
\usepackage{mathrsfs}
\usepackage{amsmath,bm}
\usepackage{cases}
\usepackage{amssymb}
\usepackage{amsthm}
\usepackage{lpic}
\usepackage{color}
\usepackage{microtype}
\usepackage{wrapfig}
\usepackage{float}
\usepackage{amsfonts}
\usepackage{eucal}
\usepackage{microtype}
\usepackage{graphicx}
\usepackage{subfigure}
\usepackage{caption}
\usepackage{url}
\usepackage{graphicx}
\usepackage{amsfonts}
\usepackage{mathtools}
\usepackage{scrextend,verbatim}
\usepackage[show]{ed}
\usepackage{tikz}

\usepackage{array, multirow}
\usepackage{tabularx} 

\usepackage{tikz}
\newcommand\checkmarkk[1][]{%
  \tikz[scale=0.4,#1]{\fill(0,.35) -- (.25,0) -- (1,.7) -- (.25,.15) -- cycle;}%
}
\newcommand\crossmark[1][]{%
  \tikz[scale=0.4,#1]{
    \fill(0,0)--(0.1,0) .. controls (0.5,0.4) .. (1,0.7)--(0.9,0.7) ..  controls (0.5,0.5) ..(0,0.1) --cycle;
    \fill(1,0.1)--(0.9,0.1) .. controls (0.5,0.3) .. (0,0.7)--(0.1,0.7) .. controls (0.5,0.4) ..(1,0.2) --cycle;
  }%
}

\usetikzlibrary{shapes,arrows,arrows.meta,positioning,fit}
\tikzstyle{block} = [rectangle, fill=white,
    minimum height=2em, node distance=12em, draw=black!50] 
\tikzstyle{line} = [draw=black!100, -{Latex[length=3mm]}]

\usepackage{tikz}
\usetikzlibrary{shapes.geometric, arrows}

\tikzstyle{startstop} = [rectangle, rounded corners, 
minimum width=3cm, 
minimum height=1cm,
text centered, 
draw=black, 
fill=red!30]

\tikzstyle{io} = [trapezium, 
trapezium stretches=true, 
trapezium left angle=70, 
trapezium right angle=110, 
minimum width=3cm, 
minimum height=1cm, text centered, 
draw=black, fill=blue!30]

\tikzstyle{process} = [rectangle, 
rectangle stretches=true, 
minimum width=3cm, 
minimum height=1cm, 
text centered, 
text width=3cm, 
draw=black, 
fill=orange!30]

\tikzstyle{decision} = [diamond, 
minimum width=3cm, 
minimum height=1cm, 
text centered, 
draw=black, 
fill=green!30]
\tikzstyle{arrow} = [thick,->,>=stealth]

\usepackage{algcompatible}

\usepackage[ruled,vlined]{algorithm2e}

\newcommand{\beq}{\begin{equation}}
\newcommand{\eeq}{\end{equation}}

\newcommand{\ed}{\end{displaymath}}
\newcommand{\bea}{\begin{eqnarray}}
\newcommand{\eea}{\end{eqnarray}}
\newcommand{\beas}{\begin{eqnarray*}}
\newcommand{\eeas}{\end{eqnarray*}}
\newcommand{\bbc}{\begin{center}}
\newcommand{\ec}{\end{center}}

\newcommand{\bfig}{\begin{figure}}

\newcommand{\efig}{\end{figure}}
\newcommand{\bverb}{\begin{verbatim}}
\newcommand{\everb}{\end{verbatim}}

\theoremstyle{remark}

\algnewcommand\algorithmicto{\textbf{to}}
\algnewcommand\RETURN{\State \textbf{return} }

\definecolor{orange}{rgb}{1,0.5,0}
\title{A hybrid numerical methodology coupling reduced order modeling and Graph Neural Networks for non-parametric geometries: applications to structural dynamics problems}

\title{A hybrid numerical methodology coupling reduced order modeling and Graph Neural Networks for non-parametric geometries: applications to structural dynamics problems}

\author{
  \underline{Victor Matray}, Faisal Amlani, David Néron \\
  Université Paris-Saclay, CentraleSupélec, ENS Paris-Saclay, CNRS, \\ LMPS - Laboratoire de Mécanique Paris-Saclay,\\
  Gif-sur-Yvette, France \\
 \textit{ Corresponding author:} Victor Matray, \texttt{victor.matray@ens-paris-saclay.fr} \\
   \And
  Frédéric Feyel \\
Université Paris-Saclay, CentraleSupélec, ENS Paris-Saclay, CNRS, \\ LMPS - Laboratoire de Mécanique Paris-Saclay,\\
  Gif-sur-Yvette, France \\
  Safran Tech, Digital Sciences \& Technologies Department, \\ Magny-les-Hameaux, France \\
  \texttt{} \\
}

\begin{document}
\maketitle

\begin{abstract}
\noindent This work introduces a new approach for accelerating the numerical analysis of time-domain partial differential equations (PDEs) governing complex physical systems. The methodology is based on a combination of a classical reduced-order modeling (ROM) framework and recently-introduced Graph Neural Networks (GNNs), {where the latter is trained on highly heterogeneous databases of varying numerical discretization sizes.} {The proposed techniques} are {shown to be} particularly {suitable} for non-parametric geometries, {ultimately enabling the treatment} of a diverse range of geometries and topologies. {Performance studies are presented in an application context related to} the design of aircraft seats and their {corresponding} mechanical responses to shocks, {where the main motivation is} to reduce the computational burden and enable the rapid design iteration {for such problems that entail non-parametric geometries}. {The methods proposed here are straightforwardly} applicable to other scientific or engineering problems {requiring a large-number of finite element-based numerical simulations}, with the potential to significantly enhance efficiency {while maintaining reasonable accuracy}.
\end{abstract}


\section{Introduction}
\label{sec:nonpara}

\noindent Numerical simulations are essential tools in a variety of scientific disciplines, including design engineering, physics, {biology,} and economics. {They} often involve resolving complex models governed by physical or phenomenological laws {that can be} mathematically described by partial differential equations (PDEs). In solid mechanics, the resolution of these models relies mainly on {mesh-based} numerical methods requiring {discretization in space and time}, the most {widely-used} of which {is} the Finite Element Method (FEM). In recent {decades}, a great deal of work {\cite{sabat_history_2021, FEM6, FEM7, FEM8, FEM5, FEM4, FEM3, FEM2, FEM1}} has been {undertaken} to improve and extend classical {FEM} formulations {in order to} encompass a wider variety of both linear and nonlinear material behavior to the extent that {FEM-based simulations} are now ubiquitous in {science and engineering}~\cite{sabat_history_2021, zhu2018finite, FEM5}.

However, despite advances in computing capacity in recent years, the computational cost of FEM-based {solvers} remains significant and often limiting for highly-sensitive {engineering or scientific} applications such as those related to aeronautical design~\cite{tzanakis_structural_2023, cavagna2011neocass, fredriksson1986advanced} {(the primary motivating context of this contribution)}. Iterating over {a} design space is necessary for engineers to develop optimized configurations~\cite{karagoz_design_2019, zhu2018light, ardila2020finite}, {and each design may require some sort of computational mechanical analysis}. Hence, it is highly advantageous and important to have very fast calculation {tools} that can enable rapid pre-dimensioning/pre-sizing of mechanical components from the initial design stage, i.e., providing adequately accurate estimations of the corresponding mechanical response and behavior {without immediately requiring expensive finite element simulations}. {Indeed, such tools could help circumvent expensive full-field analyses of modifications incurred during the design phase of a project, potentially resulting in substantial time savings during validation stages.} {Fast approximate tools can ultimately give} development teams the leeway to propose original designs {(even those that are radically different from the standard)}
{whose effectiveness can be} quickly {evaluated} without the need for computationally costly {\emph{full order modeling} using FEM.

\subsection{{Acceleration by reduced-order modeling (ROM) approaches}}

A common technique for reducing the computational burden of large-scale {FEM} simulations is known as {\emph{reduced-order modeling} (ROM)} \cite{benner_survey_2015, chinesta_short_2011, rozza_reduced_2011}. Reduced-order models exploit the mathematical separability of solution fields in a full physical model, allowing one to solve a smaller, less-costly problem {while maintaining} control over the error in approximation~\cite{maday_reduced-basis_2002, rozza_volume_2020}. A {notable class} of potent reduction techniques relies on creating a low-dimensional space using a reduced-order basis (ROB) \cite{amsallem_nonlinear_2012}. {Construction of such a space} involves an initial offline learning phase, where the high-fidelity problem {(e.g., full-order FEM)} is solved, after which appropriate snapshots are carefully selected. Subsequently, in the online phase, the original high-fidelity problem is often tackled using a projection, typically employing a Galerkin-type approach  {(although alternative methods are also feasible~\cite{carlberg2017galerkin})}. For an in-depth review of such projection-based methods, readers are encouraged to consult~\cite{benner_survey_2015} and the associated references.

{The focus of this work is on linear structural dynamics problems}, {where the} most {common ROM technique} is the {use of the} Proper Orthogonal Decomposition (POD) \cite{sirovich_turbulence_1987, POD1, POD2}. {POD} has {also} {demonstrated} superior capabilities {to} well-established modal analysis techniques {which are also widely used} \cite{radermacher_comparison_2013}. {The principle of this approach is to build a ROB from intelligently-selected snapshots taken from previously-calculated solutions.} Techniques such as the Reduced Basis Method (RBM) \cite{maday_reduced-basis_2002, rozza_volume_2020} offer automatic selection procedures to determine the most relevant snapshots. RBMs can also be used to certify the results derived from the reduced model for any parameter contained in the PDE. For a deep overview, the interested reader can refer to \cite{hesthaven_certified_2016} and the references therein. Note that ROB methods can also be extended to non-linear PDEs, e.g., via the Discrete Empirical Interpolation Method \cite{tiso_discrete_2013}.
These aforementioned {ROB} methods are generally referred to as \textit{a posteriori} methods and {their costs are} characterized by the time needed to calculate the database (offline) and the time needed to solve the reduced model (online).

In a {mathematically-related but fundamentally-different} approach, the Proper Generalised Decomposition (PGD) \cite{ladeveze_sur_1985, chinesta_short_2011, daby-seesaram_hybrid_2023} provides {an alternative technique for} low-rank resolution, but \textit{a priori}. This is accomplished by generating the ROB \emph{on-the-fly}. {Similarly to} POD, the PGD method was initially proposed for space-time variables (known as radial approximations~\cite{ladeveze_sur_1985}). {However,} PGD has since been extended to other parameters such as those defining material properties \cite{chinesta_short_2011}. Non-linear equations can also be considered with this approach, {e.g., by use of} the LATIN-PGD algorithm~\cite{ladeveze_nonlinear_1999, scanff_weakly-invasive_2022, heyberger2012multiparametric}. 

Although {reduced-order} models are becoming more increasingly used in practice~\cite{casenave_nonintrusive_2020, scanff_weakly-invasive_2022}, they are only truly effective when the problem can be parameterized (e.g., for describing material properties, loading characteristics{, or} geometries~\cite{rozza_advances_nodate}), and even more so in the context of multi-query simulations~\cite{daby-seesaram_hybrid_2023}.  Geometric considerations, in particular, often require that the number of {degrees-of-freedom} remains constant {among the offline training set} when constructing a ROB (so that the information contained in the previously-calculated snapshots can be re-evaluated). This “geometry parameterization" {restriction} is the most limiting condition in the context of design innovation and iteration since, in essence, a very original structure born of an engineer's imagination may not necessarily share the same parameterizations of other (even similar) designs.

When geometry is parametrizable (e.g., by length or width and enven more when the topologu evolves), ROM techniques are already efficient \cite{rozza_advances_nodate, lupini_use_2019}. However, their applicability is limited to geometries that can be parameterized by a few control points \cite{rozza_advances_nodate}. General techniques include Free-Form Deformation (FFD) \cite{sederberg_free-form_1986, koshakji_free_2013} as well as interpolation using Radial Basis Functions (RBF) \cite{buhmann_radial_2003, morris_cfd-based_2008}. These methods define parameters as the shifting of specific control points that determine the morphing of the domain. For instance, in \cite{lassila_parametric_2010}, airfoil profiles of the NACA family are studied using RBMs in the context of inverse design for aircraft wings. In \cite{courard_pgd-abaques_2016}, another approach is employed where a multi-parametric PGD is used to obtain numerical maps for a rocket launcher component whose geometry is parameterized with two control points that are added to the spatial and temporal parameters of the PGD.

When the geometry is non-parametric (e.g., can be of any arbitrary shape or size), strategies for ROM are difficult to apply. The geometries do not share the same spatial dimension (after FEM discretisation), making classical ROB projection methods infeasible. There are also recently-introduced approaches based on mesh morphing~ \cite{ye_data-driven_2024, casenave_mmgp_2023}, the aim of which is to learn the transformation that transfers any mesh to a reference mesh, which then enables a common spatial dimension to be obtained without a priori knowledge of the geometry parameterization. Searching for this transformation is equivalent to determining an on-the-fly parameterization of the geometries in the database, and assumes that the new geometries tested will have the same parameterization. In \cite{ye_data-driven_2024}, this technique is coupled with a ROM to simulate blood flow. Similarly, in \cite{casenave_mmgp_2023}, a similar strategy is applied to a database generated with five known material parameters and one unknown geometric parameter, which is then coupled to a Gaussian process method. Learning the morphing transformations as well as choosing reference meshes is not very straightforward \cite{casenave_mmgp_2023}. Indeed, these transformations cannot necessarily be generalized, particularly when the topology of the mesh changes \cite{alexa_recent_2002}.

{The goal of the present work is to propose a} more general approach {that requires neither parameterizing the geometry with a finite number of control points nor learning} the morphing transformation, ultimately enabling consideration of {completely unstructured} meshes {of highly-variable degrees-of-freedom}.

\subsection{{Acceleration by deep learning-based approaches}}

 An alternative class of methods that have been gaining interest for more rapidly and efficiently {approximating solutions to} PDEs is the burgeoning field of deep learning, which has now touched almost all scientific disciplines~\cite{karniadakis_physics-informed_2021, cueto2023thermodynamics}. Deep learning has shown remarkable capability in predicting non-linear relationships with physical data ~\cite{lu_deeponet_2021}. {Similarly to classical reduced-order modeling}, one needs to distinguish between two {computational cost} scales: the {\emph{offline} learning/training time (which is often quite long) and the \emph{online} model inference time (which is very short---possibly real-time)}. A great deal of {recent work} has consisted of trying to enrich deep learning models with physical knowledge \cite{karniadakis_physics-informed_2021, vadyala2022review}. To this end, there are three main types of approaches. The first, proposed in \cite{raissi_physics-informed_2019} and often referred to as “physics-informed" (or “learning bias"), consists of including physical knowledge in the loss function used to train the neural networks. The idea is to construct a loss function as the sum of a term {that accounts for deviations} from target data (error) and {a} term {that penalizes} non-physical behavior. A second approach, often referred to as “physics-augmented" (or “inductive bias"), consists in introducing physical knowledge \emph{directly} into the architecture of the neural network~\cite{hernandez2023thermodynamics}. For example, in \cite{benady_nn-mcre_2023}, the neural network is constrained to be convex in the same way as the thermodynamic potential that the authors set out to learn. Similarly, in \cite{hernandez_thermodynamics-informed_2022}, a positive dissipation is constrained to satisfy the second law of thermodynamics. A third approach, classically referred to as “observation bias" \cite{kashefi_point-cloud_2021}, aims to intelligently introduce a data structure underlain by physical knowledge. Hence, the deep neural network (DNN) is exposed directly to the observed data and is expected to capture the underlying physical process via training.

Other learning-based approaches that may ignore this physical knowledge {a priori}---instead focusing on representations adapted to physical simulation data---also show remarkable capabilities. This is the case of approaches based on Graph Neural Networks (GNN), initially proposed in \cite{battaglia_relational_2018}, which have become  very popular in several communities: in fluid mechanics for Navier-Sokes equations (e.g., \cite{li_graph_2022, pfaff_learning_2021}), in meteorology for global weather prediction (e.g., \cite{lam_graphcast_2022}), in biomechanics for describing the mechanics of human body organs (e.g., \cite{dalton_emulation_2022}), in Newtonian mechanics (e.g., \cite{bishnoi_enhancing_2022}), and in solid mechanics (e.g., \cite{deshpande_magnet_2023,GNNFEM}). Several recent works have also improved this approach by explicitly incorporating physical knowledge (see \cite{hernandez_thermodynamics-informed_2022} and references therein). {Such methods} offer interesting extrapolation capabilities and {are} particularly well-suited {for their application to a variety of geometries}~\cite{nastorg_implicit_2023} (which is an important consideration for the structural design context of this work).

 GNN and other deep learning approaches are based on training a model from a database which, in {the motivating industrial applications of the present work}, represents an opportunity to reuse the high-fidelity computations produced by other designs and engineering projects that are kept in databases. It should be noted that in this framework, the amount of available data is often limited, and this constraint is taken into account in the proof-of-concept work presented here. In addition, deep learning models (in general) face uncertainty as to the validity of their outputs which, {together with their more-often-than-not lack of interpretability \cite{karniadakis_physics-informed_2021, zhang2021survey}, often hinder their certification and {widespread} acceptance {in engineering}.  For this reason as well, the idea of combining deep learning approaches (e.g., GNN) with well-established solvers used in practice may reveal promising directions. 

\subsection{{Present work}}

This contribution proposes a new numerical methodology to provide near real-time structural mechanical analysis combining the two distinct approaches discussed above: classical ROM methods and deep learning. The general idea is to use the latter to predict the former for any arbitrary discretization size and geometric shape (in order to address the limitations of classical approaches as outlined above). More formally, we consider a set of $N$ independent geometries $\{\bm{\Omega }_p^{\text{data}}\}_{p=\{1,\dots,N\}}$ (which do not share any geometric parameter or discretisation size, as simplified in \autoref{fig:ilu_intro}} ), where each configuration is associated with a reference solution obtained by FEM: $\{\field{U }_p^{\text{data}} \}_{p=\{1,\dots,N\}}$. We then determine the associated ROBs designated by $\{\vect{P}_p^{\text{data}} \}_{p={1,\dots,N\}}}$ (for example, via PGD). In order to provide predictions for a geometry $\bm{\Omega}^{\text{new}}$ that may fall outside of the scope of treatment via the original ROBs, this work proposes the use of GNN to provide a surrogate general ROB generator that provides $\vect{P}_{\text{GNN }}$, that is trained via supervised learning on 
 $\{\vect{P }_p^{\text{data}}, \}_{p=\{1,\dots,N\}}$. The solution ${\field{U}}^\text{new}$ of the full field is then obtained by a Galerkin projection on $\vect{P}_{\text{GNN}}$, similarly to a classical ROM approach. Note that from this ${\field{U}}^\text{new}$ solution it is possible to enrich it with new PGD modes, calculated in the usual way if this is necessary to obtain a satisfactory solution.  A summary of the overall method, which we call GNN-PGD generator, is presented as a flow chart in \autoref{fig:logigrame}.

\begin{figure}[!ht]
  \center
  \noindent\includegraphics[width=12cm]{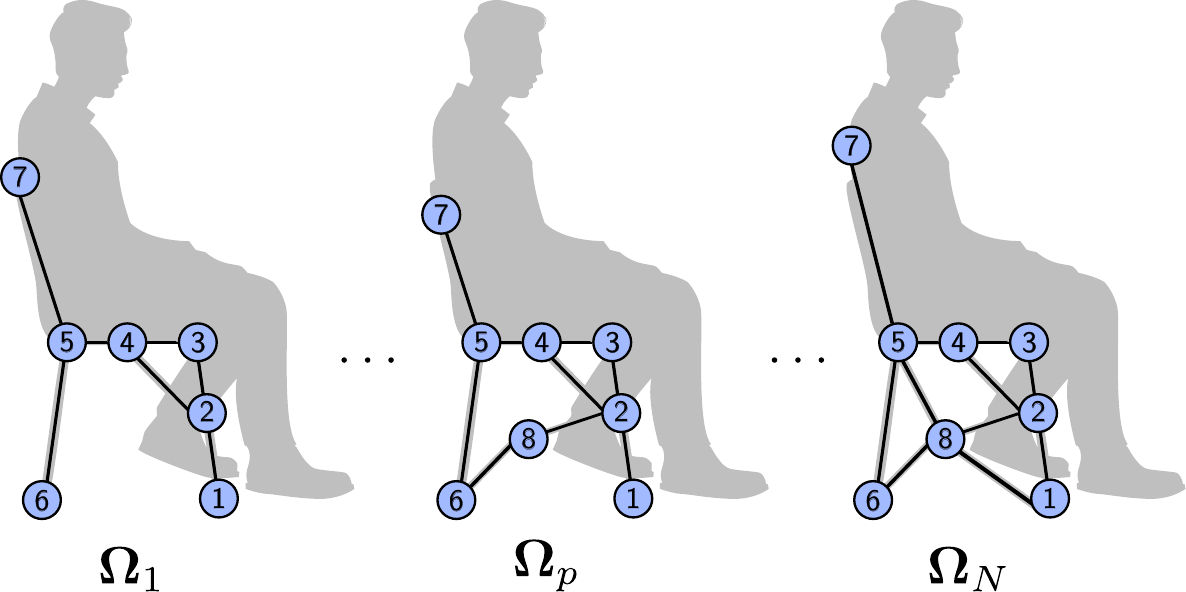}
\caption{Example illustrations of $N$ non-parameterized $\bm{\Omega}_p$ geometries with possibly different discretization sizes.}
\label{fig:ilu_intro}
\end{figure}

Again, the ultimate aim is to enable consideration of much wider and more general ranges of geometries and topologies (i.e., a ROB built on very heterogeneous databases). The algorithmic framework we propose here aims to overcome the constraints mentioned in the previous subsection of traditional ROM approaches by exploiting the flexibility and inference speed of GNNs. The manuscript is organized as follows: \autoref{sec:gouverning} describes the governing PDE of interest as well as the construction of (finite element-based) reference solutions. Thus, \autoref{sec:methodology} describes {the proposed solver} combining a {GNN} architecture with a classical ROM algorithm. Finally, {\autoref{sec:casestudy} presents a relevant proof-of-concept case study applying the new method to a problem of aircraft seat design, including performance comparisons with more conventional autoregressive GNN approaches \cite{pfaff_learning_2021}.}

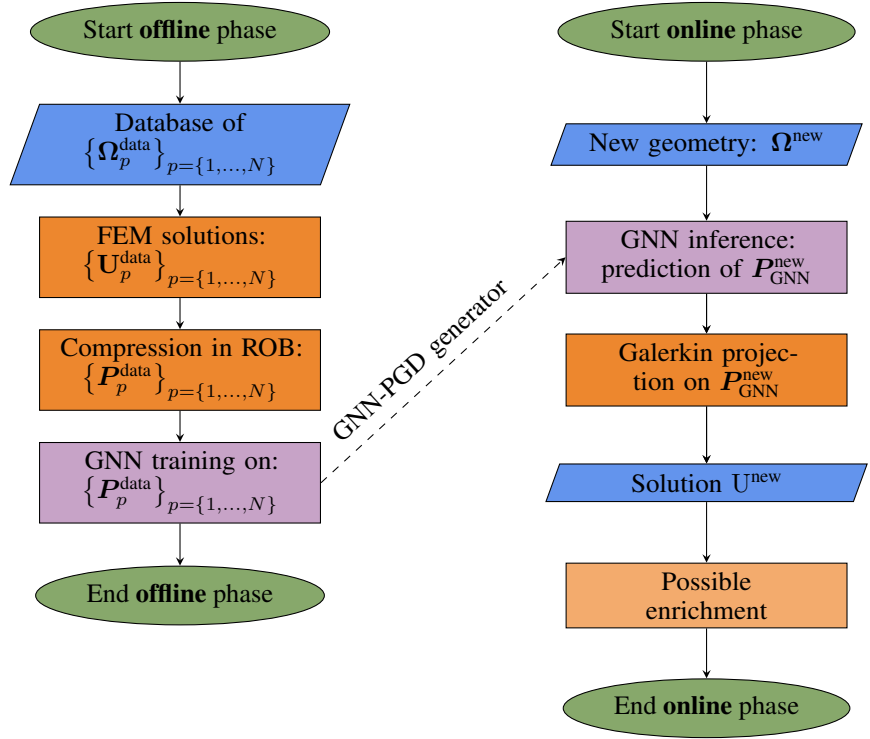
\begin{figure}[!ht]
    \centering
    \definecolor{asparagus}{rgb}{0.53, 0.66, 0.42}
\definecolor{cadmiumorange}{rgb}{0.93, 0.53, 0.18}
\definecolor{cornflowerblue}{rgb}{0.39, 0.58, 0.93}
\definecolor{lilac}{rgb}{0.78, 0.64, 0.78}

\begin{tikzpicture}[>=stealth, node distance=1.5cm, auto, on grid]

\tikzstyle{start} = [ellipse, draw, fill=asparagus, text centered, minimum width=3cm, minimum height=0.5cm]
\tikzstyle{input} = [trapezium, trapezium left angle=70, trapezium right angle=110, draw, fill=cornflowerblue, text centered, text width=3.5cm, minimum height=0.5cm]
\tikzstyle{process} = [rectangle, draw, fill=cadmiumorange, text centered, text width=3.5cm, minimum height=0.5cm]
\tikzstyle{process2} = [rectangle, draw, fill=cadmiumorange!70, text centered, text width=3.5cm, minimum height=0.5cm]
\tikzstyle{process3} = [rectangle, draw, fill=lilac, text centered, text width=3.5cm, minimum height=0.5cm]
\tikzstyle{end} = [ellipse, draw, fill=asparagus, text centered, minimum width=3cm, minimum height=0.5cm]

\node (start) [start] {Start \textbf{online} phase};
\node (design) [input, below=of start] {New geometry: $\bm{\Omega}^{\text{new}}$};
\node (gnn) [process3, below=of design] {GNN inference: prediction of $\vect{P}_{\text{GNN}}^{\text{new}}$};
\node (projection) [process, below=of gnn] {Galerkin projection on $\vect{P}_{\text{GNN}}^{\text{new}}$};
\node (solution) [input, below=of projection] {Solution ${\mathrm{U}}^{\text{new}}$};
\node (solution2) [process2, below=of solution] {Possible \\ enrichment};
\node (end) [end, below=of solution2] {End \textbf{online} phase};

\draw[->] (start) -- (design) node[midway, left, text width=0cm, align=center] {};
\draw[->] (design) -- (gnn) node[midway, left, text width=0cm, align=right] {};
\draw[->] (gnn) -- (projection) node[midway, left, text width=0cm, align=right] {};
\draw[->] (gnn) -- (projection) node[midway, right, text width=0cm, align=left] {};
\draw[->] (projection) -- (solution) node[midway, left, text width=0cm, align=right] {};
\draw[->] (projection) -- (solution) node[midway, right, text width=0cm, align=left] {};
\draw[->] (solution) -- (solution2) node[midway, left, text width=0cm, align=center] {};
\draw[->] (solution2) -- (end) node[midway, left, text width=0cm, align=center] {};

\node (start1) [start, left=7cm of start] {Start \textbf{offline} phase};
\node (bdd) [input, below= of start1, node distance=2.5cm] {Database of $\left\{\bm{\Omega}_p^{\text{data}}\right\}_{p=\{1,\dots,N\}}$};
\node (bddd) [process, below= of bdd, node distance=2.5cm] {FEM solutions: $\left\{\field{U}_p^{\text{data}}\right\}_{p=\{1,\dots,N\}}$};
\node (bdd2) [process, below= of bddd, node distance=2.5cm] {Compression in ROB: $\left\{\vect{P}_p^{\text{data}}\right\}_{p=\{1,\dots,N\}}$};
\node (gnn2) [process3, below=of bdd2, node distance=2.5cm] {GNN training on: $\left\{\vect{P}_p^{\text{data}}\right\}_{p=\{1,\dots,N\}}$};
\node (end2) [end, below=of gnn2, node distance=2.5cm] {End \textbf{offline} phase};

\draw[->] (start1) -- (bdd) node[midway, above] {};
\draw[->] (bdd) -- (bddd) node[midway, above] {};
\draw[->] (bddd) -- (bdd2) node[midway, above] {};
\draw[->] (bdd2) -- (gnn2) node[midway, above] {};
\draw[->] (gnn2) -- (end2) node[midway, above] {};

\draw[->, dashed] (gnn2.east) -- node[midway, above, sloped, text width=4cm] {\quad GNN-PGD generator} (gnn.west);

\end{tikzpicture}
    \caption{A flowchart diagram introducing the steps of the proposed GNN-PGD methodology (detailed in Section \ref{sec:methodology}).}
    \label{fig:logigrame}
\end{figure}

\newpage
\section{Governing equations}\label{sec:gouverning}

\noindent This section details the elastodynamics equations that govern our problem of interest (\autoref{subsec:PDE}) as well as the finite element method employed in this work for the construction of the full-order solutions $\left\{\field{U}_p^{\text{data}}\right\}_{p=\{1,\dots,N\}}$ associated with non-parameterized geometries $\left\{\bm{\Omega}_p^{\text{data}}\right\}_{p=\{1,\dots,N\}}$ (\autoref{subsec:FEM}). The latter is used for the supervised learning of the proposed methodology introduced in \autoref{sec:methodology}.

\subsection{Elastodynamics system}\label{subsec:PDE}

Let $\vect{x} \in \bm{\Omega}\subset\left\{\bm{\Omega}_p^{\text{data}}\right\}_{p=\{1,\dots,N\}}$, where $\bm{\Omega} \subset \mathbb{R}^{d}\,(d=1,2,3)$. We denote the boundary of $\bm{\Omega}$ by $\partial \bm{\Omega}$ which can be decomposed  into $\partial \bm{\Omega} = \partial \bm{\Omega}_{\vect{u}} \cup \partial \bm{\Omega}_{\vect{\sigma}}$, where $\partial \bm{\Omega}_{\vect{u}}$ (respectively $\partial \bm{\Omega}_{\vect{\sigma}}$) represents the parts of the domain boundary where Dirichlet (respectively Neumann) boundary conditions are imposed, with  $\partial \bm{\Omega}_{\vect{u}} \cap \partial \bm{\Omega}_{\vect{\sigma}} = \varnothing$. With a time interval of interest given by $\bm{I} = [0,T]$ for some final time $T\in\mathbb{R}^{+}$, we define the following vector spaces:

\begin{itemize}

\item[$\bullet$] $\mathcal{U}(\bm{\Omega}) = \mathcal{H}^{1}(\bm{\Omega})=\left\{\vect{u} \in \bm{L}^{2}(\bm{\Omega}), \nabla \vect{u} \in \bm{L}^{2}(\bm{\Omega})\right\}$

\item[$\bullet$] $\mathcal{U}(\bm{\Omega} ; \vect{u}_d)  = \left\{\vect{u} \in \mathcal{H}^1(\bm{\Omega}), \vect{u} \mid_{\partial \bm{\Omega}_{\vect{u}}} = \vect{u}_d  \right\}$ 

\item[$\bullet$] $\mathcal{U}(\bm{\Omega} ; \vect{0}) = \mathcal{H}^{1}_0(\bm{\Omega}) = \left\{\vect{u} \in \mathcal{H}^1(\bm{\Omega}), \vect{u} \mid_{\partial \bm{\Omega}_{\vect{u}}} = \vect{0}  \right\}$ 

\item[$\bullet$] $\mathcal{U}_h(\bm{\Omega})= \left\{\vect{u} \in \mathcal{U}(\bm{\Omega}), \vect{u}(\bm{x}) = \sum_{i=1}^{N_x} \vect{\varphi}_i (\bm{x}) u_i \right\}$

\item[$\bullet$] $\mathcal{I}_{} = \bm{L}^{2}(\bm{I}_{}, \mathbb{R})$ with norm $\|\bullet\|_{\bm{I}}=\left(\int_{\bm{I}} \bullet^{2}\mathrm{~d} t\right)^{1 / 2}$

\end{itemize}



\noindent The governing elastodynamics PDE is given by
\begin{equation}
    \label{equ}
    \rho \ddot{\vect{u}}  = \mathbf{{div}} \mat{\sigma} + \vect{f}_d, \quad (\vect{x},t) \in  \bm{\Omega} \otimes \mathcal{I}, 
\end{equation}
where $\vect{f}_d = \vect{f}_d(\bm{x},t)$ is an external volumetric force and where, for notational simplicity, we define $ \ddot{\vect{u}}={\partial^2 \vect{u}}/{\partial t^2}$ throughout this paper. Dirichlet and Neumann boundary conditions on $\partial \bm{\Omega} = \partial \bm{\Omega}_{\vect{u}} \cup \partial \bm{\Omega}_{\vect{\sigma}}$ are given by
\begin{subequations}
\begin{numcases}{}
 \bm{\vect{u}} = \vect{u}_d, & $(\vect{x},t) \in \partial \bm{\Omega}_{\vect{u}} \otimes  \mathcal{I}$     \label{BC}\\
\mat{\sigma} \cdot \vect{n} = \vect{F}_d, & $(\vect{x},t) \in \partial \bm{\Omega}_{\vect{\sigma}} \otimes  \mathcal{I}$   \label{equ2},
\end{numcases}
\end{subequations}
where $\bm{F}_d = \bm{F}_d(\bm{x},t)$ is an external surface density and where $\bm{\sigma}$ is the stress tensor given by $\mat{\sigma} = \mat{K} :  \mat{\varepsilon}$ for $(\vect{x},t) \in  \bm{\Omega} \otimes \mathcal{I}$ ($\mat{K}$ is the Hookean elastic operator and $\mat{\varepsilon}$ is the small deformation strain tensor). For all simulations in this contribution, the material is assumed to be linear elastic, homogeneous, and isotropic.
  










\subsection{Reference solution using the Finite Element Method}\label{subsec:FEM}

This section provides a brief description of the construction of the reference solutions that are used to construct the reduced databases that are used to train the GNN of Section \ref{sec:GNN}. 

\subsubsection{Weak formulation and semi-discretization in space}
\label{sec:FEM}
\noindent As in conventional FEM \cite{sabat_history_2021, FEM6, FEM7, FEM8, FEM5, FEM4, FEM3, FEM2, FEM1}, boundary conditions are imposed strongly: the unknown field $\vect{u}(\vect{x},t)$ is represented at each instant $t$ in the space of functions defined by $\bm{\Omega}$ that satisfy a priori the Dirichlet conditions of Equation~\eqref{BC}. Hence, for the weak formulation, we consider test functions $\vect{v}^* \in \mathcal{U}(\bm{\Omega}; \vect{0})$ (i.e., the space of kinematically admissible functions at zero), yielding a system for the unknown $\vect{u}(\vect{x},t)$ given by
\begin{equation}
    \label{eq:weak}
   \begin{cases}
        \operatorname{m}(\vect{v}^*, {\vect{u}}) + \operatorname{k}(\vect{v}^*, \vect{u}) = \operatorname{f}(\vect{v}^*; t), & \vect{u} \in  \mathcal{U}(\bm{\Omega} ; \vect{u}_d),  \forall \vect{v}^* \in \mathcal{U}(\bm{\Omega} ; \vect{0}), \\
        \vect{u}(\vect{x},0) = \vect{u}_0, &  (\vect{x},t) \in  \bm{\Omega} \otimes  \{0\}, \\
        \dot{\vect{u}}(\vect{x},0) = \dot{\vect{u}}_0, & (\vect{x},t) \in  \bm{\Omega} \otimes  \{0\}, \\
   \end{cases}
\end{equation}
where the scalar products $\operatorname{m}(\cdot , \cdot)$ and $\operatorname{k}(\cdot , \cdot)$ and the linear form $\operatorname{f}(\cdot ; t)$ are defined respectively by
\begin{equation}
    \label{eq:innerprod}
   \begin{cases}
         \operatorname{m}(\vect{v}^*, {\vect{u}}) = \int_{\Omega} \rho \ddot{\boldsymbol{u}} \vect{v}^* \mathrm{d} \Omega, \\
        \operatorname{k}(\vect{v}^*, \vect{u}) = \int_{\bm{\Omega}} \mat{\sigma} : \mat{\varepsilon}\left(\vect{v}^*\right) \mathrm{d}  \bm{\Omega},  \\
        \operatorname{f}(\vect{v}^*; t) = \int_{\Omega} \vect{f}_{{d}} \vect{v}^* \mathrm{d}  \Omega+\int_{\partial \Omega} \vect{F}_{{d}} \vect{v}^* \mathrm{d}  S.\\
   \end{cases}
\end{equation}

The system given by~\eqref{eq:weak} can be resolved by a Galerkin approach that searches for the solution in a subspace $\mathcal{U}_h(\bm{\Omega}) \subset \mathcal{U}(\bm{\Omega})$ of finite dimension $N_x$ with a basis given by $ \mathbf{\Phi} = [ \vect{\varphi}_1, \dots, \vect{\varphi}_{N_x} ]$. Hence an approximate solution at each time $t$ to $\vect{u}$ can be represented as $\vect{u}_h =  \mathbf{\Phi} \vectt{u}(t )$ where $\vectt{}(t) \in \mathbb{R}^{N_x}$ are the corresponding coefficients in the approximation basis $[ \vect{\varphi}_1, \dots, \vect{\varphi}_{N_x} ] $. The semi-discretized problem in space hence consists of solving the following sytem for the displacement vector $\vectt{u}: \mathcal{I} \rightarrow \mathbb{R}^{N_x}$:

\begin{equation}
\label{ODE}
\begin{cases}
\mathbb{M} \ddot{\vectt{u}}(t)+\mathbb{K}  \vectt{u}(t)  =\vectt{f}(t), \\
 \vectt{u}(0)  =\vectt{u}_0, \\
 \dot{\vectt{u}}(0)  =\vectt{v}_0,
\end{cases}
\end{equation}

\noindent where $\ope{M}=\operatorname{m}(\vect{\Phi}, \vect{\Phi})$ is the mass matrix, $\ope{K}=\operatorname{k}(\vect{\Phi}, \vect{\Phi})$ is the stiffness matrix, and $\vectt{f}(t)=\operatorname{f}(\vect{\Phi}; t)$ is a generalized force. The vectors $\vectt{u}_0 \in \mathbb{R}^{N_x}$ and $\vectt{b}_0 \in \mathbb{R}^{N_x}$ are respective cofficients of $\vect{u}_0$ and $\dot{\vect{u}}_0$ in the approximation basis $\mathbf{\Phi}$ (where, again, $\dot{\vectt{u}}={\mathrm{d} \vectt{u}}/{\mathrm{~d} t}$ and  $\ddot{\vectt{u}}={\mathrm{d}^2 \vectt{u}}/{\mathrm{~d} t^2}$).

\subsubsection{Time integration}
\label{sec:newmark}

The final ODEs in Equation~\eqref{ODE} can be integrated in time using a classical Newmark scheme \cite{newmark_method_1959} (a commonly-employed approach \cite{boucinha_spacetime_2013, daby-seesaram_hybrid_2023}). Specifically, we utilize an implicit scheme with average acceleration, which leads to a conservative and stable algorithm. The time interval of interest $\bm{I} = [0,T]$ can be discretized into $t_n = (n-1)\Delta t, n=1,...,N_t$ for $\Delta t = T/(N_t-1)$. Time is discretized uniformly with a timestep $\Delta t$. The corresponding discretized space in time (of size $N_t$), is henceforth denoted $\mathcal{I}_{\Delta t}$, and hence $\field{U} \in \mathcal{U}_h(\bm{\Omega}) \otimes \mathcal{I}_{\Delta t}$ where $\field{U}$ denotes the discrete space-time solution field.


\newpage
\section{Methodology}\label{sec:methodology}

This section details the methodology introduced in \autoref{fig:logigrame} of the introdution: the generation of the classical ROB sets $\{\vect{P }_p^{\text{data}} \}_{p=\{1,\dots,N\}}$ that are each associated with a non-parametric geometry $\left\{\bm{\Omega}_p^{\text{data}} \right\}_{p=\{1,\dots,N\}}$ (Section \ref{sec:pgd}); the proposed GNN-PGD model trained on  $\{\vect{P }_p^{\text{data}} \}$  in order to construct/predict a ROB for an unknown geometry $\bm{\Omega}_{\text{test}}$ (Section \ref{sec:GNN}); and the Galerkin projection step employed to solve the corresponding time-dependent ODE given by Equation \eqref{ODE} (Section \ref{sec:galerkin}). A summary of the overall methodology is presented in Section \ref{sec:algorithm}.

\subsection{Database generation}
\label{sec:pgd}

\noindent PGD \cite{ladeveze_sur_1985, chinesta_short_2011, daby-seesaram_hybrid_2023} is employed to compress the reference FEM solutions of Section \ref{subsec:FEM} in the form of separated variable mode products, i.e., each reduced-order basis contained in $\{\vect{P }_p^{\text{data}}\}_{p=\{1,\dots,N\}}$ corresponds to PGD spatial modes. Modes produced by such a decomposition have been shown to provide a better representation of a solution than the eigenmodes of the structure \cite{radermacher_comparison_2013}, especially for non-linear problems \cite{daby-seesaram_hybrid_2023}.

\noindent In order to explain the principles of PGD we temporarily use a continuous formulation of the referential problem of the equation \eqref{equ}. Given a known solution $\vect{u}(\vect{x},t): \bm{\Omega} \times \bm{I} \to \mathcal{U}(\bm{\Omega}) \otimes \mathcal{I}$,  the method of separating space-time variables via PGD consists of seeking an appropriate approximation $\vect{u}_M \in \mathcal{U}(\bm{\Omega}) \otimes \mathcal{I}$ of $\vect{u}(\vect{x},t)$ in the form given by \cite{ladeveze_nonlinear_1999}
\begin{equation}
    \label{eq:pgd}
    \vect{u}(\vect{x},t) \approx \vect{u}_M (\vect{x},t) = \sum_{m=1}^M \vect{\Lambda}_m(\vect{x}) \lambda_m (t),
\end{equation}
where $\vect{\Lambda}_m \in \mathcal{U}(\bm{\Omega}) $ is a spatial mode and $\lambda_m \in \mathcal{I}$ is a temporal mode (their product, $\vect{\Lambda}_m\lambda_m$, is often called a space-time mode \cite{ladeveze_nonlinear_1999}). The total number of modes $M$ is called the \emph{rank} of the approximation \cite{ladeveze_nonlinear_1999}.

The construction of mode $m$ is typically obtained through a greedy algorithm that seeks a rank $1$ approximation and minimizes its error with respect to the approximation error of rank $m-1$ \cite{chinesta_short_2011, nouy_priori_2010, ladeveze_nonlinear_1999}, i.e.,
\begin{equation}\label{eq:posteriori}
\left\{\begin{array}{l}
\vect{\Lambda}_m, {\lambda}_m=\arg \min_{ \mathcal{U}(\bm{\Omega}) \times \mathcal{I}}\left(\left\| \underbrace{\vect{u}(\vect{x},t)-\vect{u}_{m-1}(\vect{x},t)}_{\text{\tiny approximation error of rank $m-1$}}- \underbrace{\vect{\Lambda}_m(\vect{x}) {\lambda}_m(t)}_{\text{\tiny new approximation of rank 1}} \right\|^2_{\bm{L^2}} \right) \\
\vect{u}_{m-1}(\vect{x},t) =\sum_{k=1}^{m-1} \vect{\Lambda}_k(\vect{x}) {\lambda}_k(t)
\end{array}\right.,
\end{equation}

\noindent where $\|u\|_{\bm{L}^2}=\left(\int_{\mathcal{I}} \int_{\bm{\Omega}} \vect{u}^2(\vect{x}, t) \mathrm{d}\bm{\Omega} \mathrm{dt}\right)^{1 / 2}$. Defining $\vect{\Delta u}=\vect{u}-\vect{u}_{m-1}$ and exploiting  stationarity, Equation \eqref{eq:posteriori} can be solved as a coupled problem given by \cite{ladeveze_nonlinear_1999}:
\begin{equation}\label{eq:ptfixe}
\begin{cases} \forall {\lambda}^{*} \in \mathcal{I}, \int_{I} {\lambda}^{*}\left(\int_{\Omega}(\vect{\Delta u}-\vect{\Lambda}_m {\lambda}_m) \vect{\Lambda}_m \mathrm{d} \Omega\right) \mathrm{dt}=0  &  \Longleftrightarrow \quad   {\lambda}_m =\frac{\int_{\bm{\Omega}} \vect{\Delta u} \vect{\Lambda}_m d \Omega}{\int_{\bm{\omega}} \vect{\Lambda}_m^{2} d \Omega} = f_u(\vect{\Lambda}_m) \\ \forall \vect{\Lambda}^{*} \in \mathcal{U}(\bm{\Omega}) , \int_{\Omega} \vect{\Lambda}^{*}\left(\int_{I}(\vect{\Delta u}-\vect{\Lambda}_m {\lambda}_m) {\lambda}_m \mathrm{dt}\right) \mathrm{d} \Omega=0 & \Longleftrightarrow \quad \vect{\Lambda}_m =\frac{\int_{I} \vect{\Delta u} {\lambda}_m \mathrm{dt}}{\int_{I} {\lambda}_m^{2} \mathrm{dt}}=g_u({\lambda}_m) \end{cases}
\end{equation}

\noindent The solutions $\lambda_m, \vect{\Lambda}_m$ exist and are unique up to a multiplicative factor. In order to ensure overall uniqueness of the solutions in \eqref{eq:ptfixe}, the spatial mode is usually normalized \cite{ladeveze_nonlinear_1999}. A fixed-point algorithm, using stagnation of the temporal mode as a stopping criterion, can be used to find such solutions with guaranteed convergence since seeking the best rank $r$ approximation with a given $\vect{\Delta u}$ can be shown to be equivalent to a generalized eigenvalue problem \cite{nouy_priori_2010}.

In practice, the final rank $M$ is selected based on an error $\epsilon$ that serves as the stopping criterion for the greedy algorithm, i.e.,

\begin{equation}\label{eq:citere}
\frac{\left\| \vect{u}-\bm{u}_M\right\|_{\bm{L}^2}^{2}}{\left\| \vect{u}\right\|_{\bm{L^2}}^{2}} \leqslant \epsilon.
\end{equation}
This ultimately enables automatic selection of the optimal number of modes for a given $\epsilon$, which can be computationally advantageous over SVD-based approaches that require calculation of the complete set of modes of the original field $\vect{u}$ \cite{boucinha_spacetime_2013} in order to form an approximation or basis.

In the discrete sense (where the FEM-based reference solution is given by $\field{U} \in \mathcal{U}_h(\bm{\Omega}) \otimes \mathcal{I}_{\Delta t}$), one seeks the separated basis above in the corresponding finite-dimentional  space, i.e., $ \forall m, \vect{\Lambda}_m \in \mathcal{U}_h(\bm{\Omega}), \vect{\lambda}_m \in \mathcal{I}_{\Delta t}$, and hence the reduced-order approximation is given by
\begin{equation}\label{eq:reducedU}
\field{U} \simeq \field{U}_M=\sum_{m=1}^M \vect{\Lambda}_m \otimes \vect{\lambda}_m.
\end{equation}
Although $N\times N_t$ real numbers are needed to store the full field $\field{U}$, the representation for $\field{U}_M$ in Equation \eqref{eq:reducedU} requires only $M\times(N+N_t)$ real numbers for $M<<N$.

Hence for each element of the set of arbitrary geometries $\left\{\bm{\Omega}_p^{\text{data}} \right\}_{p=\{1,\dots,N\}}$, the associated ROB (via a PGD on their corresponding reference solutions as described above) can be given by
\begin{equation}\label{eq:ROB}
\bm{P}_p^{\text{data}} = \operatorname{Span} \left( \left\{\vect{\Lambda}_{p,m}^{\text{data}}\right\}_{m=\{1,\dots,M\}} \right),
\end{equation}
\noindent where $\bullet^{\text{data}}$ denotes the labeled data for a training database. The set of each ROB, $\{\vect{P }_p^{\text{data}}, \}_{p=\{1,\dots,N\}}$, constitutes the complete database for the supervised learning employed in the GNN component (detailed in the following subsection) of the GNN-PGD generator introduced in this work.
\newpage

\subsection{Deep learning on graphs: Graph Neural Networks (GNNs)}\label{sec:GNN}
The following presents an overview of GNNs and additionally details the implementation of our proposed architecture. As a reminder, the aim of the GNN-PGD generator is to propose a ROB to describe the solution field of a new, arbitrary, geometry $\bm{\Omega}_{\text{test}}$. In order to accomplish this, the GNN-PGD generator is trained on the database ($\{\vect{P }_p^{\text{data}}\}_{p=\{1,\dots,N\}}$) presented in the previous subsection. \autoref{fig:conv_gnn} provides an illustrative summary of the main steps of a GNN.

\subsubsection{Overview of GNN}

\noindent The formalism of graph neural networks (GNNs) offers a powerful approach for data processing of numerical problems involving meshes. A mesh can be interpreted as a graph comprising nodes $i$ for $i=1,...,n$ linked by edges. These nodes and edges contain features $\vect{x_}i$ and $\vect{e}_{ij}$ respectively. The graph also contains mesh connectivity information in the form of an adjacency matrix $\mat{A}$ (which is simply a reformulation of the connectivity table deriving from an FEM problem). As an example, the adjacency matrix of the graph in \autoref{fig:conv_gnn} is given by
\begin{equation*}
\small
\mat{A} = \left( \begin{matrix}
0 & 1 & 0 & 0 & 0 & 0 & 0 \\
1 & 0 & 1 & 1 & 0 & 0 & 0 \\
0 & 1 & 0 & 1 & 0 & 0 & 0 \\
0 & 1 & 1 & 0 & 0 & 0 & 0 \\
0 & 0 & 0 & 1 & 0 & 1 & 1 \\
0 & 0 & 0 & 0 & 1 & 0 & 0 \\
0 & 0 & 0 & 0 & 1 & 0 & 0 \\
\end{matrix} \right)_{n \times n}
\end{equation*}
In this work, we consider undirected graphs, which result in symmetrical $\mat{A}$ adjacency matrices. Moreover, we don't add any self-loops to the nodes~\cite{hamilton_graph_nodate}, which enables a diagonal of zeros in $\mat{A}$.

\begin{figure}[!ht]
  \center
  \noindent\includegraphics[width=15cm]{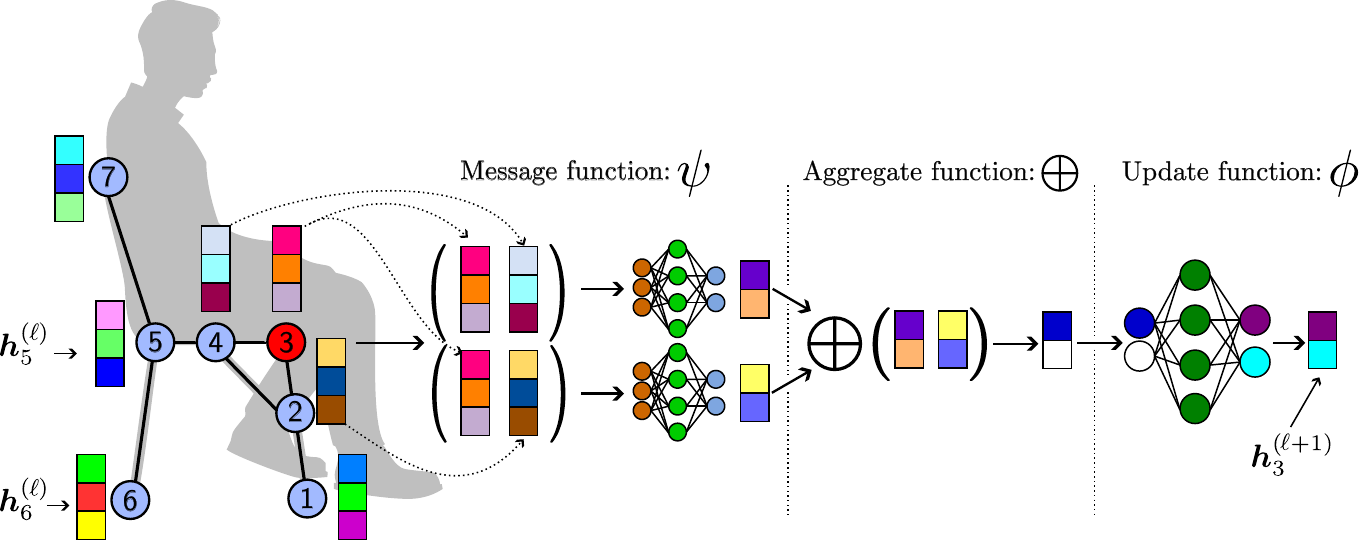}
\caption{Illustration of a Message-Passing layer on a three-stage graph. Neural networks symbolize functions whose parameters can be adjusted by learning (here applied to node three of the example graph of $n=7$ nodes).}
\label{fig:conv_gnn}
\end{figure}

GNNs are recognized as a generalization of Convolutional Neural Networks (CNNs) applied to graph-structured data. Traditionally, CNNs have been used for Cartesian grid-structured data such as images \cite{lecun_convolutional_2010}. This generalization enables GNNs to detect and understand complex structures in data by applying convolution-like operations, making them extremely effective for, e.g.,  regression tasks.

Our proposed architecture uses the Message-Passing GNN (MP-GNN) formalism \cite{hamilton_graph_nodate}, which is a mathematical formulation local to the node level. In MP-GNN models, the latent feature vector $\vect{h}_i$ of a given node $i$ is derived by employing a permutation-invariant aggregator function $\bigoplus$, such as a sum or a mean, over the feature vectors of its neighbors $\mathcal{N}(i)$. Additionally, each neighbor's feature vector may undergo transformation by a function $\psi$. Subsequently,  $\bigoplus$ may undergo further transformation by another function $\phi$. This sequence of operations constitutes a GNN layer\cite{hamilton_graph_nodate}. Formally, the equation governing the feature vector $\vect{h}_i^{(l+1)}$ of a node $i$ in the subsequent GNN layer $\ell+1$ can be expressed as
\begin{equation}
\label{gnnlayer}
\vect{h}_i^{(\ell+1)}=\phi\left(\vect{h}_i^{\ell}, \bigoplus_{j \in \mathcal{N}(i)} \psi\left(\vect{h}_i^{\ell}, \vect{h}_j^{\ell}\right)\right),
\end{equation}
which is illustrated in \autoref{fig:conv_gnn}.

Each layer $\ell$ is thus defined by the three functions $(\bigoplus, \psi, \phi)$. The message transmission function $\psi$ and the update function $\phi$ are learnable and enable us to exhibit the important graph characteristics required for learning. The aggregation function $\bigoplus$ is \emph{not} learnable and must be permutation invariant. This latter characteristic ensures that the order and number of neighboring nodes has no impact on the dimension of the aggregation result. This flexibility enables GNN to handle graphs of varying sizes, making them a powerful tool for regression applications on graph-structured data, and therefore potentially on physical problems requiring meshing of any arbitrary size. For further details, the reader is reffered to~\cite{besta2024parallel}.

\subsubsection{GNN-PGD implementation and training strategy}
\label{sec:gnndetails}

 The most commonly-found approach for physical problems \cite{dalton_emulation_2022, li_graph_2022, lam_graphcast_2022, bishnoi_enhancing_2022}, originally proposed in \cite{pfaff_learning_2021}, is known as \textsc{MeshGraphNet}. Such a method employs an encode-process-decode architecture of \cite{battaglia_relational_2018}, which consists of several multilayer perceptrons (MLPs) \cite{rosenblatt_perceptron_1958} shared between all nodes of the graph. This architecture is invoked to design our GNN-PGD algorithm, which is summarized in \autoref{algo:GNNPGD} (described with a local perspective, at the scale of node $i$, for clarity, an ddetailed in this section). In addition to the connectivity matrix $\mat{A}$ of the graph, the input to the GNN-PGD procedure contains $k$ characteristics (to be considered) of each mode $i$, i.e., $\vect{x}_i \in  \mathbb{R}^k$ (forming a row of a matrix we call $\mat{X}$). These characteristics are specific to the $\bm{\Omega}$ geometry and can include, for example, the position of the nodes or the type of boundary conditions to which they are subjected (Dirichlet or Neumann). This is clarified with further details that are presented in \autoref{sec:casestudy}.
 
 \begin{algorithm}[!ht]
  \caption{Details of the GNN implementation}
  \SetKwInOut{Input}{inputs}
  \SetKwInOut{Output}{output}
  \SetKwProg{GNNPGD}{GNN-PGD}{}{}
  \SetKwProg{Encode}{Encode}{}{}
  \SetKwProg{Process}{Process}{}{}
  \SetKwProg{Decode}{Decode}{}{}

  \GNNPGD{$(\mat{X}, \mat{A})$}{
    \Input{$\vect{x}_i$ input feature vector of node $i$ \\ $\mat{A}$ graph adjency matrix}
    \Output{$\vect{y}_i$ output feature vector of node $i$}
    
    \Encode{}{
    $\vect{h}_i^{0}=\operatorname{MLP}_{\text{node\_encoder}}(\vect{x}_i)$ \\
   
    }
    \Process{}{
    \For{$0 \leqslant \ell \leqslant L-1$}{
    $\vect{h}_i^{(\ell+1)}=\phi^{\ell}\left(\vect{h}_i^{\ell}, \bigoplus_{j \in \mathcal{N}(i)} \psi^{\ell}\left(\vect{h}_i^{\ell}, \vect{h}_j^{\ell}, \underbrace{\vect{x}_i - \vect{x}_j}_{\text{skip-connections}} \right)\right)$ \\
    Instance Normalisation of $\vect{h}_i^{(\ell+1)}$
    }
    }
    \Decode{}{
    $\vect{p}_i=\operatorname{MLP}_{\text{node\_decoder}}(\vect{h}_i^{L})$
    }
    \KwRet{$\vect{p}_i$}\;
  }
  \caption{GNN-PGD description}
  \label{algo:GNNPGD}
\end{algorithm}
 
For a general output ${\vect{P}} = [\vect{p}_1,...,\vect{p}_n]^{\top}$, the goal is to predict an image of the ROB associated with the geometry $\bm{\Omega}$, i.e., at each node $i$, we predict the output $\vect{p}_i \in \mathbb{R}^g$. Since the ROB is associated with $M$ spatial modes, and since GNN provides results at the node level, each node has $g=M\times d$ features (with $d=1,2,3$, the dimension of the problem). In other words, $\vect{p}_i$ contains the projections of the ROB ${\vect{P}}$ along the different $d$ dimensions of the $\bm{\Omega}$ geometry. In the training phase, this output ${\vect{P}}$ is compared to the given labeled training data $\left\{\vect{P}_p^{\text{data}}\right\}_{p=\{1,\dots,N\}}$ in order to adjust the weights of the GNN-PGD generator. In the inference phase (online phase), we designate this output by $\vect{P}_{\text{GNN}}^{\text{new}}$. A summary of the notation and nomenclature for the GNN-PGD proposed in this work is given in \autoref{tab:gnn}.
 
 \begin{table}[!ht]

\centering
\begin{tabular}{ll}
\hline & \multicolumn{1}{c}{ \textbf{Structure of graph inputs} } \\
\hline$G=(V, E)$ & A graph; $V$ and $E$ are sets of vertices and edges, \\
$n, m$ & Numbers of vertices and edges in $G ;|V|=n,|E|=m$, \\
$\mathcal{N}(i)$ & 1-hop neighborhood of $i$, \\
$\mat{A} \in \mathbb{R}^{n \times n}$ & The graph adjacency matrices. \\

\hline & \multicolumn{1}{c}{ \textbf{Structure of GNN computations} } \\
\hline
$L$, & The number of GNN layers, \\
$k$, & The number of input node features,\\
$g$, & The number of output node features,\\
$\vect{X} \in \mathbb{R}^{n \times k}$, & Input node features matrix, \\
${\vect{P}} \in \mathbb{R}^{n \times g}$, & Outputs node features matrix, \\
$\vect{x}_i, \vect{p}_i, \vect{h}_i^{\ell}$, & Input, output, and hidden feature vector of a node $i$ (layer  $\ell$ ). \\
\hline
\end{tabular}
\caption{Nomenclature of symbols used to describe the proposed GNN-PGD algorithm.}\label{tab:gnn}
\end{table}

The three steps in~\autoref{algo:GNNPGD} are as follows. In the first step (\textbf{Encode}), an MLP is utilized to transform the input quantities $\vect{x}_i$ into a latent space of dimension $H$.  In this work, we employ a two-layer MLP with SiLU activation functions \cite{elfwing_sigmoid-weighted_2017} and a normalization layer on the output \cite{wu_group_2018}. The second step (\textbf{Process}) employs the GNN layers. Details of the MLP architecture used here are given in table \ref{tab:detailsGNN}. In addition to the information provided by previous layers, the corresponding message functions $\psi^{(l)}$ take into account the non-transformed physical quantities $\vect{x}_i$ (known as skip-connections, as proposed in DenseNet \cite{huang_densely_2018}). After each convolution, an instance normalization layer \cite{ulyanov_instance_2017} is applied in order to improve convergence and stability of the learning phase without a dependence on the batch size used during learning \cite{wu_group_2018} (unlike batch normalization layers \cite{ioffe_batch_2015}). Our proposed method employs the same input and output dimensions (of size $H$) during this process phase, thanks to our choice of MLP architecture and the use of a permutation-invariant aggregation function. The final third step (\textbf{Decode}) in~\autoref{algo:GNNPGD} transforms the process output from the latent dimension $H$ to the physical output dimension $g$. In order to achieve this, we use another two-layer MLP (with SiLU as the activation function for the first layer) and a linear activation function for the output layer. The architectures described above are summarized in~\autoref{tab:detailsGNN}.

\begin{table}[htbp]
    \centering
    \renewcommand{\arraystretch}{1.5}
    \begin{tabularx}{\linewidth}{|>{\centering \arraybackslash}m{3cm}|>{\centering\arraybackslash}m{1.2cm}|>{\centering\arraybackslash}m{1.2cm}|>{\centering\arraybackslash}X|>{\centering\arraybackslash}m{2cm}|}
        \hline
        \rowcolor{gray!50}
        Function & Inputs Size & Outputs Size & Details & Learnable? \\
        \hline
        {$\operatorname{MLP}_{\text{node\_encoder}}(\bullet)$}  & k & H & linear ; SiLU ; linear ; SiLU ; LayerNorm & \checkmarkk[green] \\
        
        \hline
        $\psi^{\ell}(\bullet)$ & H & H & linear ; SiLU ; linear ; SiLU & \checkmarkk[green]\\
        \hline
        $\bigoplus(\bullet)$ & H & H & mean function & \crossmark[red, scale=1]\\
        \hline
        $\phi^{\ell}(\bullet)$ & H & H & linear ; SiLU ; linear ; SiLU  & \checkmarkk[green] \\
        \hline
        $\operatorname{MLP}_{\text{decoder}}(\bullet)$ & H & q & linear ; SiLU ; linear   & \checkmarkk[green] \\
        \hline
    \end{tabularx}
    \caption{Description of the architectures of the various MLPs employed to build the GNN-PGD algorithm.}
    \label{tab:detailsGNN}
\end{table}

For each bath $b$, the loss function employed in this work for the supervised training of the GNNs is a Mean Squared Error (MSE) function that is weighted by hyperparameters $\left\{\eta_m\right\}_{m=\{1,\dots,M\}}$ (each associated with a spatial mode that comprises the ROB $\vect{P}$) and is given by
\begin{equation}
\mathcal{L}_b=\sum_{m=1}^{M} \eta_m \operatorname{MSE}\left(\vect{\Lambda}_{b,m}-\vect{\Lambda}_{b,m}^{\mathrm{data}}\right),
\end{equation}
where $\vect{\Lambda}_{b,m}$ is the $m$-th mode associated with the ROB $\vect{P}$ of batch $b$, and where $\vect{\Lambda}_{b,m}^{\mathrm{data}}$ is the $m$-th mode associated with the data $\vect{P}_b^{\text{data}}$. Hence the total loss is simply given by
\begin{equation}\label{eq:loss}
\mathcal{L}=\sum_{b=1}^{N_{\text{batch}}} \mathcal{L}_b.
\end{equation}
Using this loss function,  optimization for training proceeds with a stochastic gradient descent algorithm AdamW \cite{loshchilov_decoupled_2019}, which incorporates $\bm{L}^2$ regularization on the GNN weights to prevent overfitting. This algorithm is controlled by two hyperparameters: the learning rate $\alpha$ and the ''weight decay'' $\gamma$ (for regularization). A noise percentage of $5\%$ of the average value of $\mat{X}$ is added to the inputs of the GNN-PGD in order to improve the regularisation of the network. \cite{santos2022avoiding}.

A generalized summary of the GNN-PGD hyperparameters employed in this work is provided in~\autoref{tab:hyper}. These hyperparameters can significantly influence the quality of learning and, consequently, the performance of the final trained GNN-PGD model. In order to optimize these hyperparameters---typically an exploratory and experimental task \cite{casenave_mmgp_2023, brandstetter_message_2022}---we propose the use of a specific method based on a systematic design of experiments (DOE) that is characterized by two types of variables: the maximum number of models to be tested ($100$ in all subsequent results of this paper), and the intervals of values that the aforementioned hyperparameters are permitted to take in each model (discrete or continuous). The $100$ models are generated using the DOE method called maximum projection Latin Hypercube Sampling (LHS), which enables an efficient exploration of all the different possible model combinations~\cite{joseph_maximum_2015}. The general idea is to train these 100 models of different values of hyperparameters for a reasonable number of epochs and then select the most relevant model based on a common metric. The models identified as most relevant or accurate are then re-trained for a larger number of epochs.

\begin{table}[!ht]
\centering
\begin{tabular}{ll}
\hline & \multicolumn{1}{c}{ \textbf{Hyperparameters} } \\
\hline $H$ & dimension of the latent space, \\
$L$ & numbers of GNN layers, \\
$\eta_m$ & weight of learned modes, \\
$\alpha$ & the learning rate, \\
$ \gamma$ & the weight decay, \\
$b_s$ & the batch size. \\
\end{tabular}
\caption{Hyperparameters that are optimized for the GNN-PGD learning using a Design of Experiments (DOE) procedure known as Latin Hypercube Sampling (LHS)~\cite{joseph_maximum_2015}.}
\label{tab:hyper}
\end{table}

At the end of the offline phase (see \autoref{fig:logigrame}) described in this section, a GNN-PGD generator is obtained, which constructs a ROB $\vect{P}_{\text{GNN}}=\operatorname{GNN-PGD} (\bm{\Omega}^{\text{new}})$ for a new (arbitrary) geometry $\bm{\Omega}^{\text{new}}$. As described in the following section, this ROB can then be used to solve the associated ROM for the complete problem.

\subsection{Galerkin projection on the trained ROB}
\label{sec:galerkin}

To complete the solver, Equation~\eqref{ODE} for $n\times d$ degrees-of-freedom is solved in the form of a product of functions with separated variables, similar to common methods of projection onto reduced bases \cite{radermacher_comparison_2013, rozza_reduced_2011, benner_survey_2015}, i.e.,
\begin{equation}\label{eq:forme_sol}
{\vectt{u}}^{\text{new}}(t)=\vect{P}_{\text{GNN}}^{\text{new}}{\boldsymbol{\lambda}(t)},
\end{equation}
where $\vect{P}_{\text{GNN}}^{\text{new}}$ is the ROB in space obtained by GNN as described in the previous section. In order to obtain the corresponding temporal coefficients of the general solution given by Equation~\eqref{eq:forme_sol}, we utilize a standard Galerkin-type projection approach~\cite{radermacher_comparison_2013} which results in an ODE-system given by
\begin{equation}\label{eq:galerkin}
{\vect{P}_{\text{GNN}}^{\text{new}}}^{\top} \cdot \mathbb{M} \, \underbrace{ \vect{P}_{\text{GNN}}^{\text{new}} \ddot{{\boldsymbol{\lambda} }}(t)}_{\ddot{{\vectt{{u}}}}^{\text{new}}(t)} +  {\vect{P}_{\text{GNN}}^{\text{new}}}^{\top} \cdot \mathbb{K} \, \underbrace{\vect{P}_{\text{GNN}}^{\text{new}} {\boldsymbol{\lambda}(t)} }_{{\vectt{{u}}}^{\text{new}}(t)}  =  {\vect{P}_{\text{GNN}}^{\text{new}}}^{\top} \cdot \vectt{f}(t),
\end{equation}
\noindent which is a system of $M$ ODEs (for each spatial mode $i =1,...M$) for the corresponding $M$ temporal unknowns $\boldsymbol{\lambda}(t) = (\lambda_1(t),...,\lambda_M(t))^{\top}$. The cost of solving this linear system is very low, since the size of the operators to be inverted is $M \ll n\times d$. Once the unknown $\lambda_i(t)$ are obtained from solving Equation \eqref{eq:galerkin}, the complete space-time solution can be reconstructed via Equation~\eqref{eq:forme_sol} (or, more explicitly, Equation \eqref{eq:pgd}).

For noise that may result in the $\vect{P}_{\text{GNN}}^{\text{new}}$ prediction, what follows is a proposed procedure describing how one can account for such errors.  Indeed, one can reasonably expect a prediction of the GNN-PGD generator that is not identical to the reference $\vect{P}$. In the next subsection, we present a noise-reducing strategy and apply it to a simulated example of artificial Gaussian noise to demonstrate its efficacy.

\subsubsection{Noise effect and resolution strategy}

In order to investigate and ultimately mitigate the impact that noise, which may be present in spatial modes, can have on the quality of the projection and resolution of \eqref{eq:galerkin}, we consider the ``true" (or ``exact") modes of the PGD decomposition (i.e., those that are constructed for the database given by $\left\{\vect{\Lambda}_m^{\text{data}}\right\}_{m=\{1,\dots,M\}}$). Gaussian noise of different levels is applied to all degrees-of-freedom of the structure, and the intensity is adjusted by a percentage of the maximum value of the modes amplitude. The corresponding ``noisy" modes are denoted $\left\{\vect{\Lambda}_m^{\text{noisy}}\right\}_{m=\{1,\dots,M\}}$ and constitute a new reduced basis given by
\begin{equation}\label{eq:projecteur_noisy}
\bm{P}_{\text{noisy}} = \operatorname{Span}\left( \left\{\vect{\Lambda}_m^{\text{noisy}}\right\}_{m=\{1,\dots,M\}}  \right).
\end{equation}

\autoref{fig:precond} presents the angle formed between the two vector spaces $\bm{P}$ and $\bm{P}_{\text{noisy}}$ as a function of percentage of artificial noise added for an example ROB solution comprised of three modes. It can be observed that for low levels of noise, these two vector spaces are very close. However, for noise levels close to $100 \%$, these two vector spaces are almost orthogonal. 

\begin{figure}[!ht]
  \center
  \noindent\includegraphics[width=11cm]{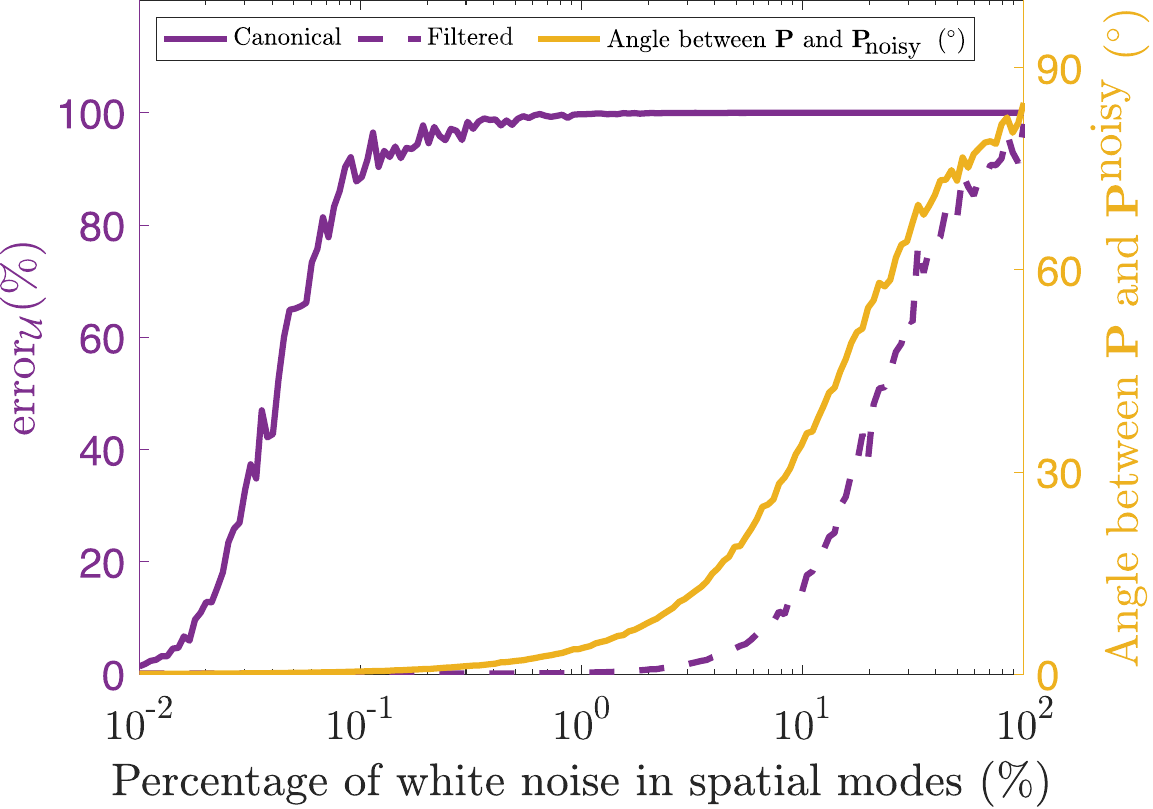}
  \caption{Effect of the presence of noise in the ROB on the resolution of the projected equation. Three modes have been used to generate the ROB.}
  \label{fig:precond}
\end{figure}

\autoref{fig:precond} additionally presents the evolution of the $\bm{L}^2$ error in the field ${\field{U}^{\text{new}}}$ (relative to the reference solution $\field{U_{\text{ex}}}$) as a function of the noise added to $\bm{P}_{\text{noisy}}$, defined by the expression
\begin{equation*}
\begin{aligned}
\label{eq:error}
\text{error}_{\mathcal{U}} &= 
 \frac{ \lVert \field{U_{\text{ex}}} - {\field{U}^{\text{new}}} \rVert_{\mathcal{U}} } {\lVert \field{U_{\text{ex}}}  \rVert_{\mathcal{U}}}, \% \quad 
\|\bullet\|_{\mathcal{U}} &= \left(\int_{\Omega \times \mathcal{I}} \bullet^2 \mathrm{~d} \Omega \mathrm{d} t\right)^{1 / 2}.
\end{aligned}
\end{equation*}
Two cases are presented: a \emph{canonical} case and a \emph{filtered} case. The former corresponds to the evolution of the error solving the original form of the ODE system (given by Equation~\eqref{eq:galerkin}), i.e., 
\begin{equation}\label{eq:canonical}
\bm{P}^{\top}_{\text{noisy}} \cdot \mathbb{M} \, \underbrace{\bm{P}_{\text{noisy}} \ddot{{\boldsymbol{\lambda} }}(t)}_{\ddot{{\vectt{{u}}}}^{\text{noisy}}(t)} +  \bm{P}_{\text{noisy}}^{\top} \cdot \mathbb{K} \, \underbrace{\bm{P}_{\text{noisy}} {\boldsymbol{\lambda}(t)} }_{{\vectt{{u}}}^{\text{noisy}}(t)}  =  \bm{P}_{\text{noisy}}^{\top} \cdot \vectt{f}(t).
\end{equation}

The \emph{filtered} curve corresponds to the error when solving a preconditioned formulation of \eqref{eq:galerkin} which, in general, is given by
\begin{equation}\label{eq:filtering}
\bm{P}^{\top}_{\text{noisy}} \cdot \mathbb{S} \mathbb{M} \, \underbrace{\bm{P}_{\text{noisy}} \ddot{{\boldsymbol{\lambda} }}(t)}_{\ddot{{\vectt{{u}}}}^{\text{noisy}}(t)} +  \bm{P}_{\text{noisy}}^{\top} \cdot \mathbb{S} \mathbb{K} \, \underbrace{\bm{P}_{\text{noisy}} {\boldsymbol{\lambda}(t)} }_{{\vectt{{u}}}^{\text{noisy}}(t)}  =  \bm{P}_{\text{noisy}}^{\top} \cdot \mathbb{S} \vectt{f}(t),
\end{equation}
where $\mathbb{S} \in \mathbb{R}^{(n\times d)^2}$ can be interpreted as a filter or a preconditioner. Here, a preconditioner of  $\mathbb{S}=\mathbb{K}^{-1} $ has been utilized, inspired by \cite{roux_optimal_2020}, where a preconditioned norm with operator $\mathbb{S}=\mathbb{K}^{-1}$ is proposed (referred to as the reconditioned equilibrium gap (REG)). In \cite{roux_optimal_2020}, the concept of ``spectral sensitivity" is introduced in order to quantify the sensitivity of different norms, constructed with different operators (such as the stiffness operator $\mathbb{K}$), towards the presence of noise from displacement field measurements (the context of that study is the identification of nonlinear behavior laws from experimental measurements by image correlation). Indeed, the $\mathbb{K}$ operator has a high spectral sensitivity~\cite{roux_optimal_2020}, reflecting its high sensitivity to noise, unlike the $\mathbb{M}$ operator. Such sensitivity can be attributed to the differential nature of $\mathbb{K}$ (see Equations \eqref{eq:innerprod} and \eqref{ODE}). On the other hand, it's the integrating nature of $\mathbb{S}$ that can correct the undesirable effects of such noise.  Choices other than $\mathbb{S}=\mathbb{K}^{-1}$  can accomplish the same objective, as long as they contain the same regularization properties of $\mathbb{K}^{-1}$. For the preliminary (linear) solver proposed here,  we employ $\mathbb{S}=\mathbb{K}^{-1}$ as a proof-of-concept: despite the prohibitive cost of calculating $\mathbb{K}^{-1}$ in a linear elastodynamics context, such cost remains negligible for nonlinear problems \cite{roux_optimal_2020} (which is the long-term objective of the methodology presented in this paper).

\autoref{fig:precond} illustrates the significant improvement that can be attained by using the filtering/preconditioning strategy (Equation \eqref{eq:filtering}) versus solving the original canonical ODE system (Equation \eqref{eq:canonical}). Indeed, for all levels of noise in $\vect{P}_{\text{noisy}}$,  the former results in significantly lower error; this implies that solving the unfiltered system of Equation \eqref{eq:canonical} is very sensitive to the noise present in $\vect{P}_{\text{noisy}}$ and may possibly lead to a very polluted ROB that has poor approximation capabilities of the original FEM system of interest. For all numerical experiments of \autoref{sec:casestudy}, the preconditioned system given by Equation \eqref{eq:filtering} is employed to obtain the temporal coefficients $\lambda_i(t)$.

\subsection{Summary of our methodology}\label{sec:algorithm}
\autoref{algo:methodo} presents a summary of the overall methodology (additionally summarized in the introductory \autoref{fig:logigrame}).
\begin{algorithm}[!ht]
  \caption{Methodology summary}
  \SetKwInOut{Input}{inputs}
  \SetKwInOut{Output}{output}
  \SetKwProg{offline}{Offline phase}{}{}
  \SetKwProg{compress}{Database compression}{}{}
  \SetKwProg{training}{GNN-PGD training}{}{}
  \SetKwProg{online}{Online phase}{}{}
  \SetKwProg{GNNPGD}{GNN-PGD inference}{}{}
  \SetKwProg{projection}{Galerkin projection}{}{}

  \SetKwProg{online}{Online phase}{}{}
   \offline{$\left(\left\{\bm{\Omega}_p^{\text{data}}, \field{U}^{\text{data}}_p \right\}_{p=\{1,\dots,N\}}\right)$}{
     \vspace{0.2cm}
        \Input{$\left\{\bm{\Omega}_p^{\text{data}}, \field{U}^{\text{data}}_p \right\}_{p=\{1,\dots,N\}}$, known geometries and associated high-fidelity solution fields, }
        \Output{$\operatorname{GNN-PGD \, generator}$: trained GNN model, }
        \compress{}{
                    \For{$0 \leqslant p \leqslant N$}{
                                        creation of the ROB $\vect{P}_p^{\text{data}}$ from the PGD compression of the high fidelity field $\field{U}_p^{\text{data}}$,
                                                }
                \KwRet{$\left\{\bm{P}_p^{\text{data}} \right\}_{p=\{1,\dots,N\}}$}
                   }
        \training{}{Supervised learning of GNN-PGD on $\left\{\bm{P}_p^{\text{data}} \right\}_{p=\{1,\dots,N\}}$ 
        
        \KwRet{GNN-PGD generator}
        }
        \KwRet{GNN-PGD generator}
   }
   
   \vspace{0.5cm}
   
   \online{$\left(\bm{\Omega}^{\text{new}}, \mathbb{M}, \mathbb{K}, \vectt{f}, \mathbb{S}\right)$}{
        \vspace{0.2cm}
        \Input{$\bm{\Omega}^{\text{new}}, \mathbb{M}, \mathbb{K}, \vect{F}, \mathbb{S}$, new geometry (unknown to GNN-PGD generator and not parameterized) and its associated operators }
        \Output{$\field{U}^{\text{new}}$: full displacement field solution }
        
        \GNNPGD{}{
        $\vect{P}_{\text{GNN}}^{\text{new}}=\operatorname{GNN-PGD} (\bm{\Omega}^{\text{new}})$: ROB proposal by GNN inference, 
        
        \KwRet{$\vect{P}_{\text{GNN}}^{\text{new}}$}
        }
        
        \projection{}{
        ${\vect{P}_{\text{GNN}}^{\text{new}}}^{\top} \cdot \mathbb{S} \mathbb{M} \, \vect{P}_{\text{GNN}}^{\text{new}} \ddot{\lambda}(t) +  {\vect{P}_{\text{GNN}}^{\text{new}}}^{\top} \cdot \mathbb{S} \mathbb{K} \, \vect{P}_{\text{GNN}}^{\text{new}}
        \lambda(t) =  {\vect{P}_{\text{GNN}}^{\text{new}}}^{\top} \cdot \mathbb{S} \vectt{f}(t)$
        
        smart Galerkin projection on $\vect{P}_{\text{GNN}}^{\text{new}}$, and resolution of the system ODE,
        
        \KwRet{$\lambda(t)$}
        
        }
        ${\vectt{u}}(t)=\vect{P}_{\text{GNN}}^{\text{new}}\lambda(t)$: full displacement field
        
        \KwRet{$\field{U}^{\text{new}}$}
        }
        \label{algo:methodo}
\end{algorithm}

\newpage 
\section{Performance/case study: simulating aircraft seat response to crash loads}\label{sec:casestudy}

\noindent In order to assess the performance of our proposed solver, we consider a case study relevant to structural design, in particular towards the problem of presizing/predimensioning of aircraft parts subjected to sudden loadings. The aircraft industry uses the acceleration (or deceleration) loading profile depicted in \autoref{fig:loading} in order to assess the strength of aircraft seats according to regulation 49 CFR Part 572, Subpart B \cite{tzanakis_structural_2023}. This loading simulates the deceleration experienced by an aircraft in the event of a collision with an obstacle. The sizing criteria focus on both the mechanical integrity of the seat components as well as the physical safety of the passenger. In the proof-of-concept methodology presented here, passenger modeling is not taken into account, and the analysis focuses solely on structural integrity. The acceleration load illustrated in \autoref{fig:loading} is applied as an inertial force at the seat structure level and also as a loading force at the seating surface (equivalent to the deceleration force experienced by a 90-kg passenger). These two contributions are reflected in the loading term $\vectt{f}(t)$ in the right-hand-side of Equation \eqref{ODE}.

\begin{figure}[!ht]
  \center
  \noindent\includegraphics[width=7cm]{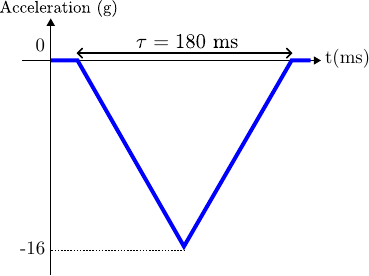}
  \caption{Loading of interest}
\label{fig:loading}
\end{figure}

\subsection{Database generation}

\noindent A database is generated using finite element simulations for the training of the GNN-PGD. In order to accomodate the common restrictions of industrial applications, we use a reasonably small number of geometries: $500$, similarly to \cite{casenave_mmgp_2023}. Out of these $500$ airplane seats, $80\%$ ($N=400$ seats) are chosen to constitute the training set (i.e., the seats seen by the GNN-PGD for adjusting parameter weights through learning), i.e., $\{\bm{\Omega}_p^{\text{data}}, \}_{p=\{1,\dots,N=400\}}$ different geometries in the training database. Another $10\%$ ($50$ seats) of the original 500-seat database is used for the validation set, representing seats unknown to the network but used to fine-tune the learning process in order to prevent overfitting by employing early-stopping criteria \cite{rice2020overfitting}. The remaining $10\%$ ($50$ seats) make up the test set on which the GNN performance is evaluated after training, representing new $\bm{\Omega}^{\text{new}}$ geometries that are completely unknown to the GNN-PGD model (see \autoref{algo:methodo}). We note that the original high-fidelity space-time fields $\left\{\field{U}_p^{\text{data}}  \right\}_{p=\{1,\dots,500\}}$, whose solutions are facilitated by FEM, have very similar separability properties. Indeed, by setting a stopping criterion of the PGD mode generation algorithm to $\varepsilon = 10^{-3}$ in \autoref{eq:citere}, we obtain a decomposition rank of $M=3$ for all set of seats, and hence train with and keep only $M=3$ modes in the ROB. A summary of the database generation procedure is provided in \autoref{fig:bdd_gene}.

\begin{figure}[!ht]
  \center
  {
  \noindent\includegraphics[width=15cm]{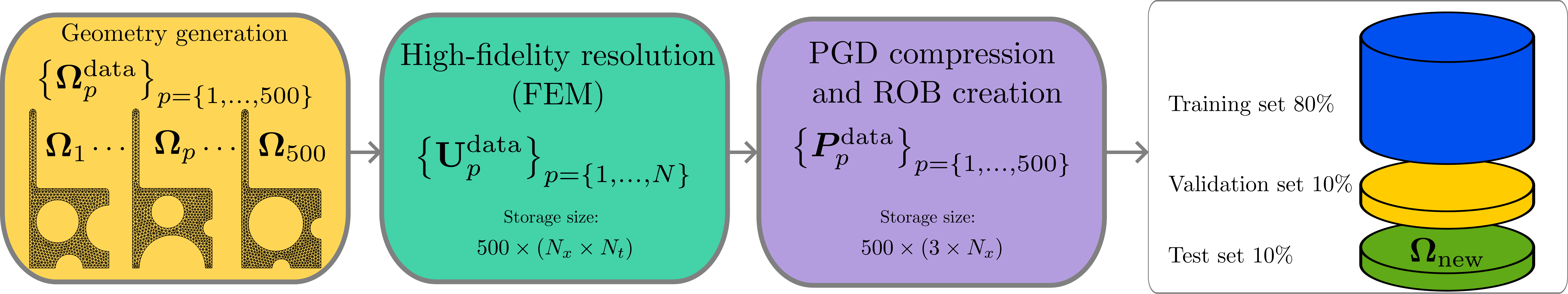}}
  \caption{Database generation }
\label{fig:bdd_gene}
\end{figure}

For the proof-of-concept presented here, we utilize a 2D modeling approach with the assumption of plane deformations (extention of the overall methodology to three-dimensions is straightforward). The material considered is Aluminum 6082-T6, commonly used in the aerospace industry \cite{tzanakis_structural_2023}, and is further assumed to be homogeneous and linear elastic isotropic. In order to generate the database from which the seats in \autoref{fig:bdd_gamma} are extracted, we employ five geometric parameters: the radii of each of the curved holes and the ($x,y$) position of the interior hole's center. Although these parameters are crucial for database generation, we recall that the solver proposed here is for non-parametric geometries, and hence are not considered beyond database generation (similarly to other approaches \cite{casenave_mmgp_2023, ye_data-driven_2024}). Indeed, such parameters are specifically unknown to the GNN-PGD model for the motivations of this paper (cf. \autoref{sec:nonpara}).
The meshes are generated by the open-source software package \textsc{GMSH} \cite{geuzaine_gmsh_2009} with T3 linear elements. In our context, we assume that the input meshes possess good quality in terms of element aspect ratios and node densities, which are adapted to the regularity of the fields of interest. The temporal scheme utilized to solve problem \eqref{ODE} is the Newmark scheme \cite{newmark_method_1959}, particularly specifically the implicit formulation with average acceleration. A temporal convergence study results in an a choice of an optimal number of time steps corresponding to $N_t=1000$. The finite element solver for generating the high-fideleity database (described in \autoref{sec:FEM} and \autoref{sec:newmark}) is self-implemented in an in-house MATLAB code.

\begin{figure}[!ht]
  \center
  {
  \noindent\includegraphics[height=7cm]{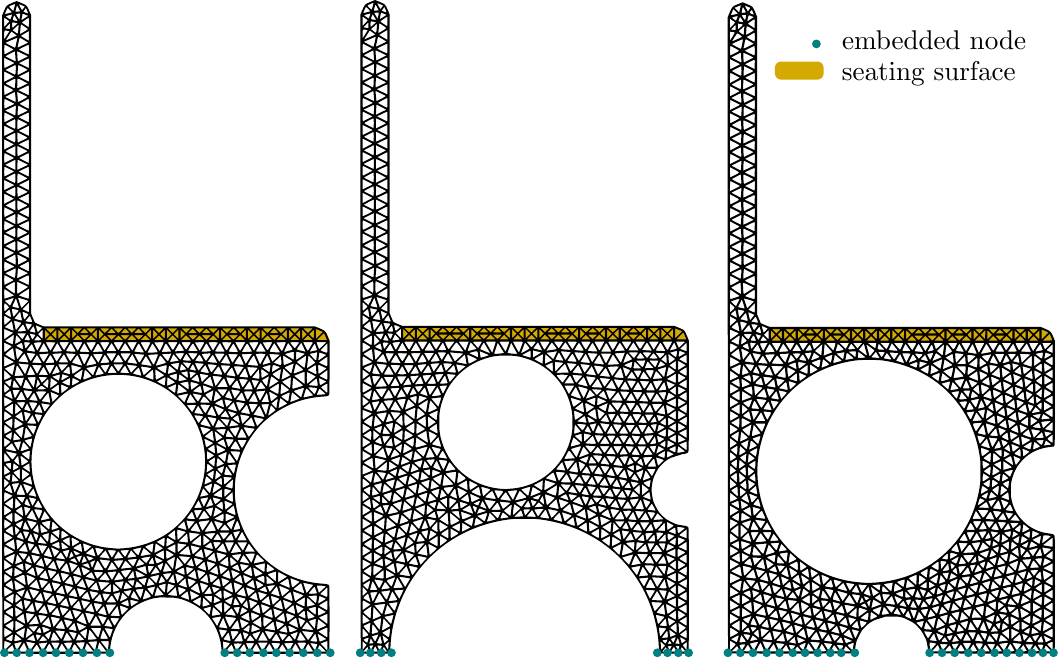}}
  \caption{Example of three seats from the database to illustrate the embedded nodes and seating surface.}
\label{fig:bdd_gamma}
\end{figure}

\subsubsection{Database characteristics}

\autoref{fig:bdd_diversite} presents database statistics in the form of histograms. It can be seen that the number of degrees-of-freedom varies considerably from seat to seat, with a total database standard deviation of $58$ degrees-of-freedom and an amplitude of $275$ degrees-of-freedom between the two extrema. Recall that these differences in degrees-of-freedom render conventional ROM approaches as impossibly to apply on heterogeneous databases such as the one considered here.

\begin{figure}[!ht]
  \centering

  \subfigure[Histogram of the distribution of the number of degrees of freedom for the 500 database seats generated.]{\includegraphics[width=5cm]{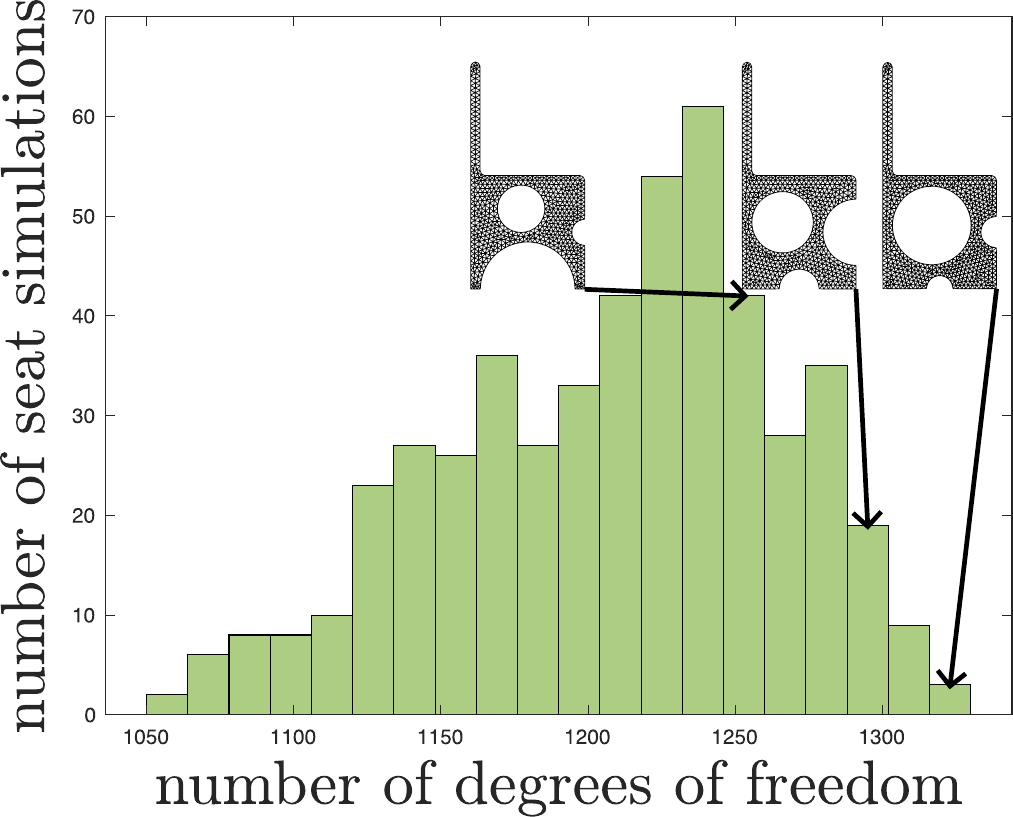}}
  \hspace{0.5cm}
  \subfigure[Histogram of the distribution of the maximum simulated displacement of the 500 seats in the generated database.]{\includegraphics[width=5cm]{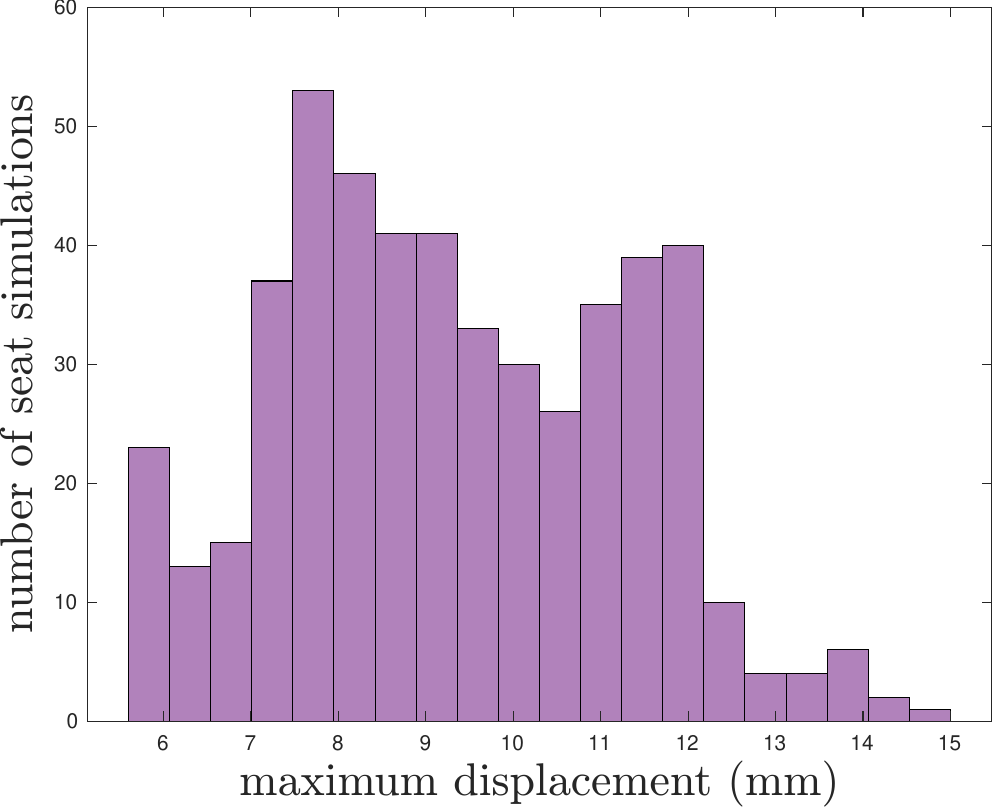}}
  \hspace{0.5cm}
  \subfigure[Histogram of the distribution of the simulated maximum Von Mises stress for the 500 seats in the generated database.]{\includegraphics[width=5cm]{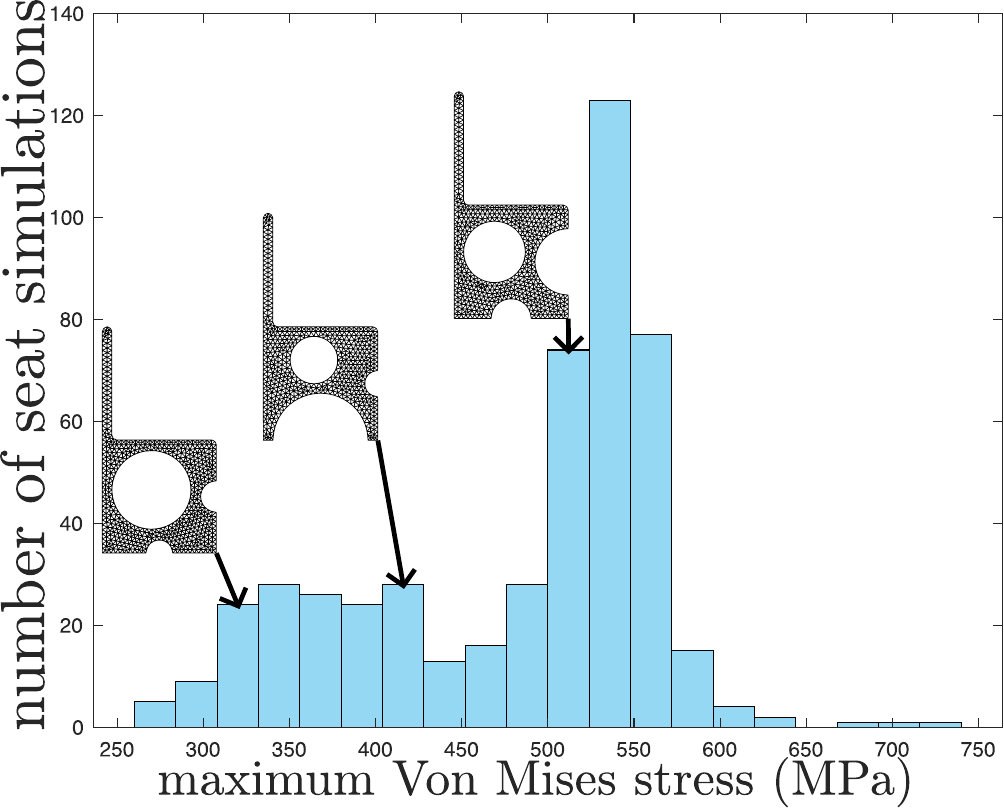}}
  \caption{Statistical analysis of the database used for training, validation, and testing.}
  \label{fig:bdd_diversite}
\end{figure}

The middle and right panels of \autoref{fig:bdd_diversite} also highlight the diversity in the overall mechanical behavior of the seats based on displacement and maximum Von Mises stress, respectively. Furthermore, the distribution of seats is clearly non-Gaussian. This variability poses a significant challenge for learning and understanding the underlying patterns in seat mechanics.

\subsection{GNN-PGD implementation details and parameters}

This section details the specific parameters employed in this case study for the GNN-PGD model introduced in \autoref{sec:gnndetails}. All algorithms were implemented with the pytorch library \cite{NEURIPS2019_9015} using pytorch-geometric \cite{Fey:2019wv}. The input matrix $\vect{X} \in \mathbb{R}^{n \times k}$ (\autoref{tab:gnn}) is composed of information on the coordinates of the node (two  components, $x$ and $y$), the type of node (free or embedded, as a boolean), the maximum-in-time of the generalized effort $\vectt{f}(t)$ at node $i$ (two spatial components, $f_x$ and $f_y$). Hence at each node, there are $k=5$ elements of input. Recall that the generic output of the GNN-PGD generator is ${\vect{P}} \in \mathbb{R}^{n \times g}$; hence the projection of the $M=3$ PGD modes here associated with each seat yields $g=3\times2=6$.

For improving learning performance of our GNN-PGD method, the input data is also pre-processed as follows. The input quantities ($\vect{X}$) are normalized according to the mean and standard deviation statistics of the training set, and the output quantities (${\vect{P}}$) are standardized according to the  maximum and minimum of the training set. 

\begin{table}[!ht]
\centering
\begin{tabular}{|c|c|c|c|c|c|c|c|}
\hline
\cellcolor{gray!50} \textbf{Hyperparameters} & $H$ & $L$ & $\eta_1$ & $\eta_2$ & $\eta_3$ & $\alpha$ & $\gamma$ \\
\hline
\cellcolor{gray!50} \textbf{Values} & 24 & 15 & 10 & 1 & 100 & $3 \times 10^{-3}$ & $5 \times 10^{-3}$ \\
\hline
\end{tabular}
\caption{The final hyperparameters employed for the results of this section.}
\label{tab:hyper2}
\end{table}

 The hyperparameters determined for these test cases are produced by a DOE (as described in \autoref{sec:gnndetails}) for 100 different models (i.e., with different hyperparameters on each) over 700 epochs. 
The table \ref{tab:hyper2} shows the resulting hyperparameters associated with the best model, which is then utilitized for training over a larger number of epochs until an early learning stopping criterion is reached. The final GNN-PGD generator has a number of $65\,022$ learnable parameters.

\autoref{fig:loss} presents the corresponding evolution of the loss as a function of epochs for the corresponding GNN-PGD model during the training phase (see \autoref{eq:loss}). We note that the GNN-PGD model updates its parameters as many times as there are batches in the training set (400 in our case). The training has been carried out on 4 Nvidia V100 GPUs and lasted 4 hours and 25 minutes (including hyperparameter optimization).

\begin{figure}[!ht]
  \center
  \noindent\includegraphics[width=11cm]{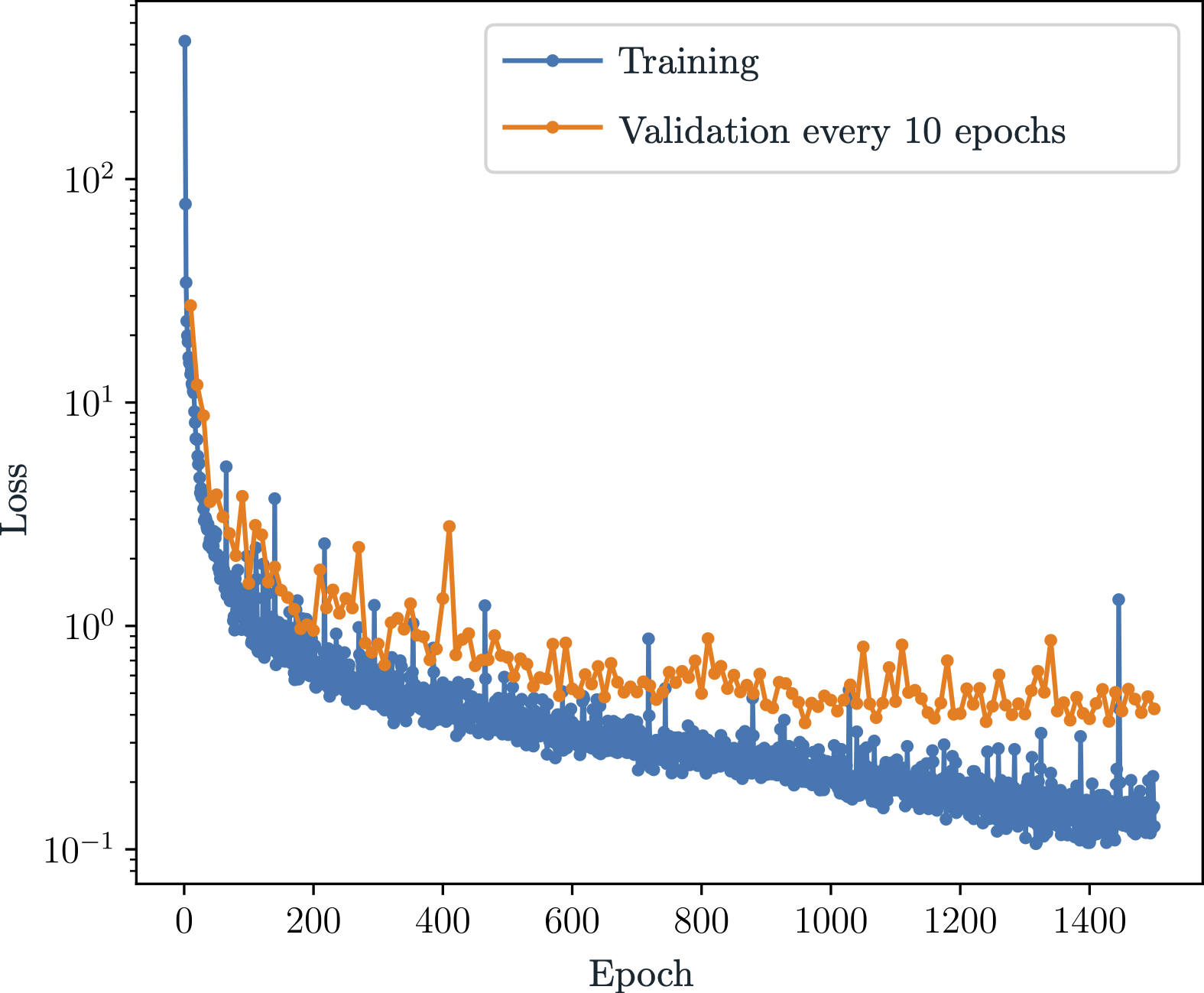}
  \caption{Evolution of the loss function of the GNN-PGD generator associated with the hyperparameters of \autoref{tab:hyper2}.}
  \label{fig:loss}
\end{figure}

\subsection{GNN-PGD learning results}

We analyze the results of the GNN-PGD training by evaluating its performance on the $50$ seats in the test set  $\left\{\bm{\Omega}_p^{\text{test}}\right\}_{p=\{1,\dots,50\}}$, i.e., seats unknown to and unseen by the GNN-PGD. 
The GNN-PGD's objective is to construct the $\left\{\vect{P}_p^{\text{test}}\right\}_{p=\{1,\dots,50\}}$ ROBs of the $50$ as best as it can. As described earlier, these ROBs are composed of $M=3$ PGD spatial modes (to be predicted by the GNN-PGD). We quantify the error between the GNN-PGD's inference prediction $\left\{\vect{P}_p^{\text{GNN}}\right\}_{p=\{1,\dots, 50\}}$ and the labeled ROBs $\left\{\vect{P}_p^{\text{test}}\right\}_{p=\{1,\dots,50\}}$ obtained from the reference field decomposition. The metric used is the normalized root-mean-square error (RMSE) for each of the $M = 3$ modes predicted, and the results are presented in \autoref{fig:rmse}.

\begin{figure}[!ht]
   
  \centering

  \subfigure[Normalized Root Mean Squared Error (RMSE) histogram for Mode 1 of the 50 tested seats.]{\includegraphics[width=5cm]{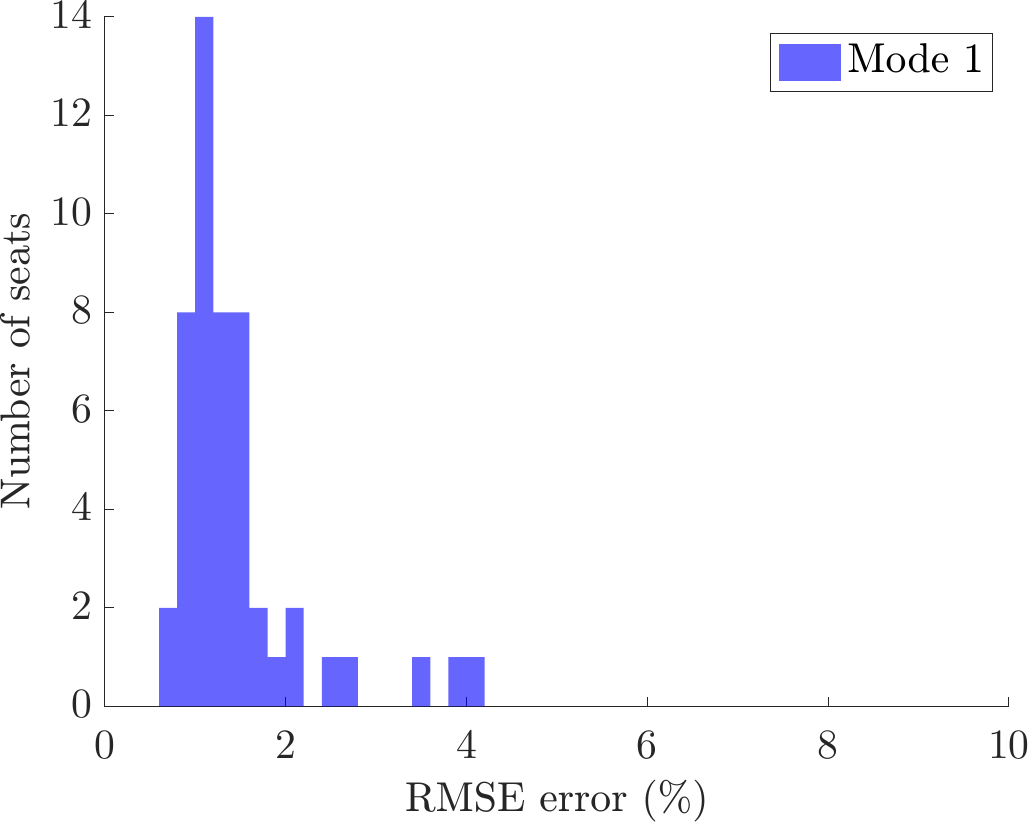}}
  \hspace{0.5cm}
  \subfigure[Normalized Root Mean Squared Error (RMSE) histogram for Mode 2 of the 50 tested seats.]{\includegraphics[width=5cm]{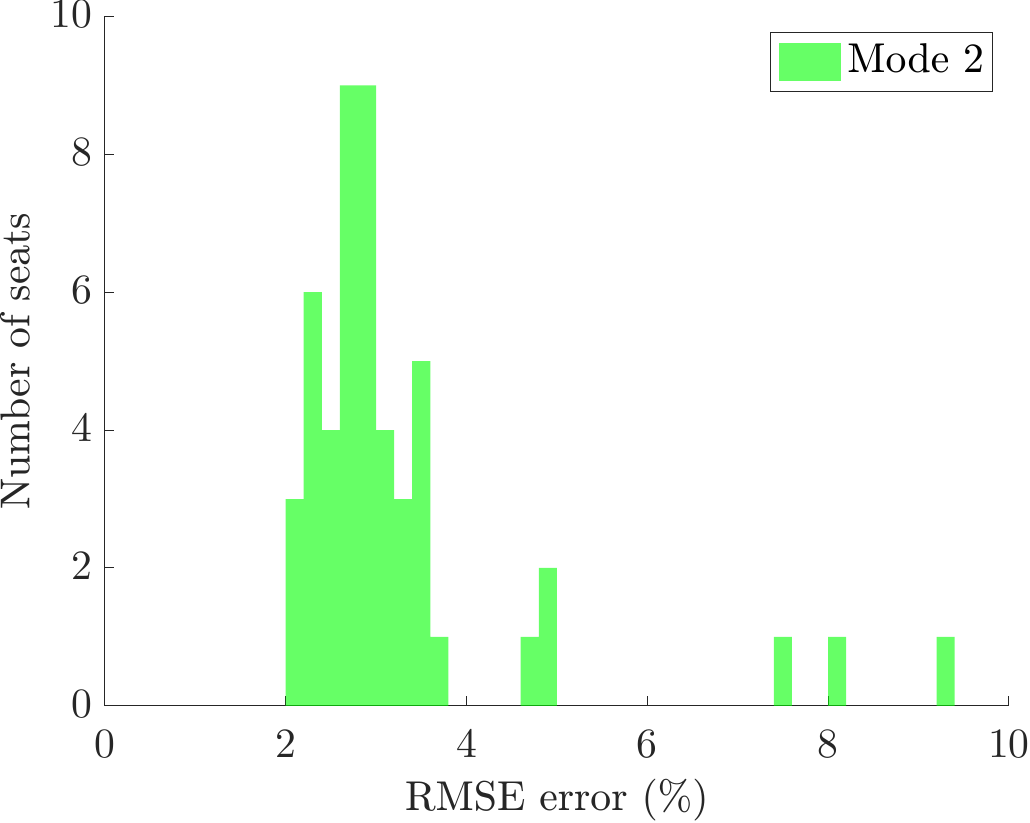}}
  \hspace{0.5cm}
  \subfigure[Normalized Root Mean Squared Error (RMSE) histogram for Mode 3 of the 50 tested seats.]{\includegraphics[width=5cm]{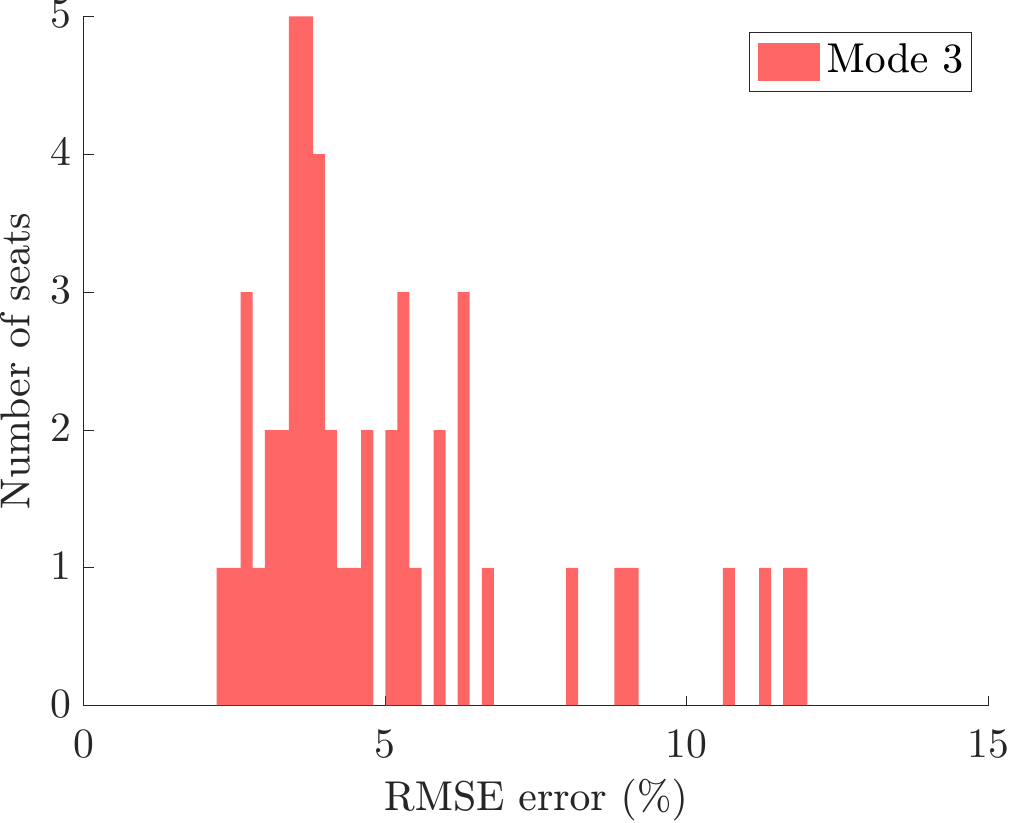}}

  \caption{Results on the modes learned by the GNN-PGD.}
   \label{fig:rmse}

\end{figure}

Results demonstrate the ability of GNN-PGD to predict with good accuracy the ROBs associated with the unknown (test) seat geometries. Indeed, errors on predicted modes are small: less than $5\%$ for Mode 1, less than $10\%$ Mode 2, and less than $15\%$ for Mode 3. We note that, as evidenced in \autoref{fig:rmse}, the errors increase on average with the rank of the predicted mode. An analysis of the seats that lie on the higher end of the error plots yields no discernible pattern to the particular geometries that have slightly higher errors. A visual comparison with the corresponding true reference modes (``Ground Truth"), including a plot of the difference, of the first, second and third spatial mode solutions predicted by GNN-PGD are provided in Figures \ref{fig:mode1_example},  \ref{fig:mode2_example}, and \ref{fig:mode3_example}, respectively. These figures demonstrate the ability of the proposed GNN-PGD solver to very accurately capture both coarse and fine behavior found in the very different modes. \autoref{fig:mode33_example} additionally presents a similar comparison for Mode 3 on a diffferent geometry than \autoref{fig:mode3_example}, further illustrating the large variations that can be found in the various modes between geometries, and the ability of GNN-PGD to accurately capture and predict such disparate behavior between different seats.  The results in general indicate the ability of GNN-PGD to predict well a ROB for a new geometry, which can then be used to obtain the complete associated spatio-temporal field via Galerkin projection (\autoref{sec:galerkin}).

\begin{figure}[!ht]
  \center
  \noindent\includegraphics[width=16cm]{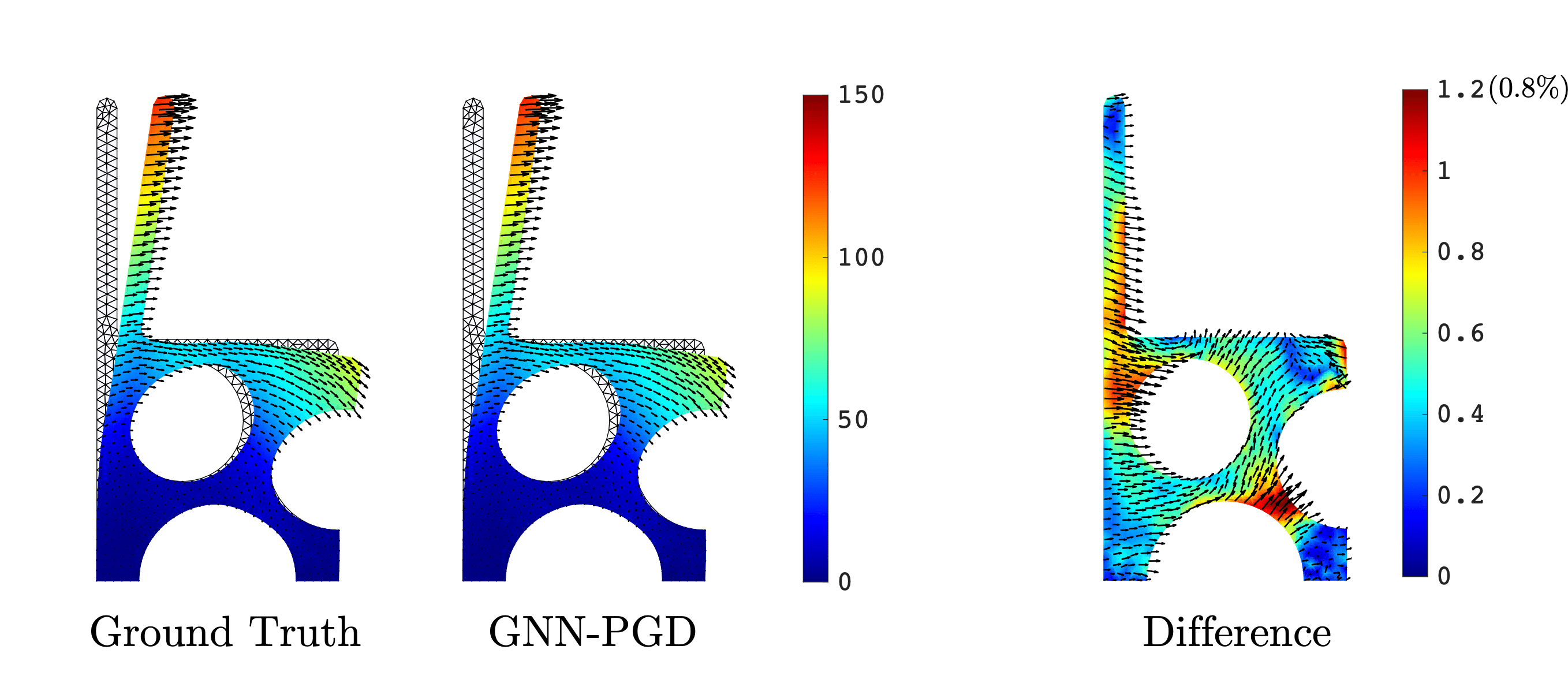}
  \caption{Example of the first mode predicted by GNN-PGD for seat 1 from the test database, together with the ground truth and error.}
  \label{fig:mode1_example}
\end{figure}

\begin{figure}[!ht]
  \center
  \noindent\includegraphics[width=16cm]{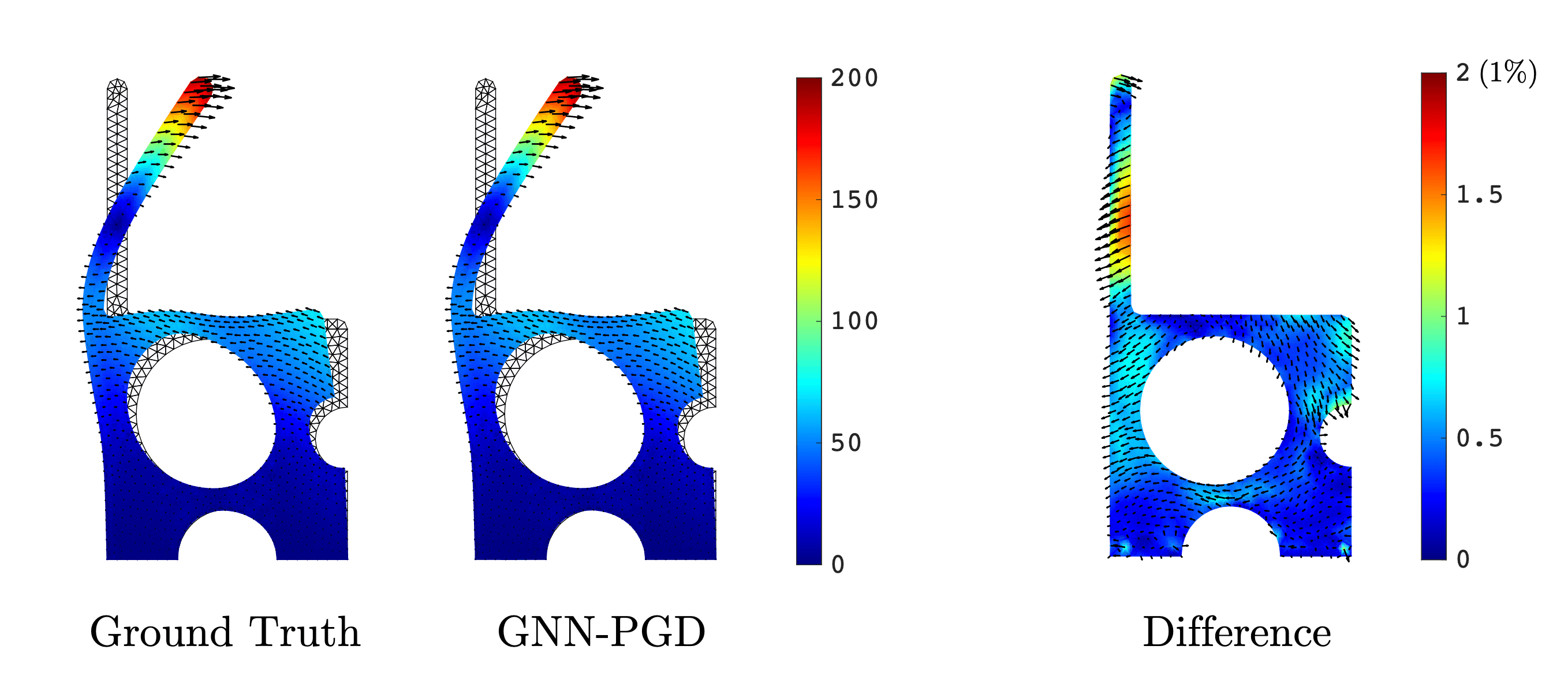}
  \caption{Example of the second mode predicted by GNN-PGD for seat 10 from the test database, together with the ground truth and error.}
  \label{fig:mode2_example}
\end{figure}

\begin{figure}[!ht]
  \center
  \noindent\includegraphics[width=16cm]{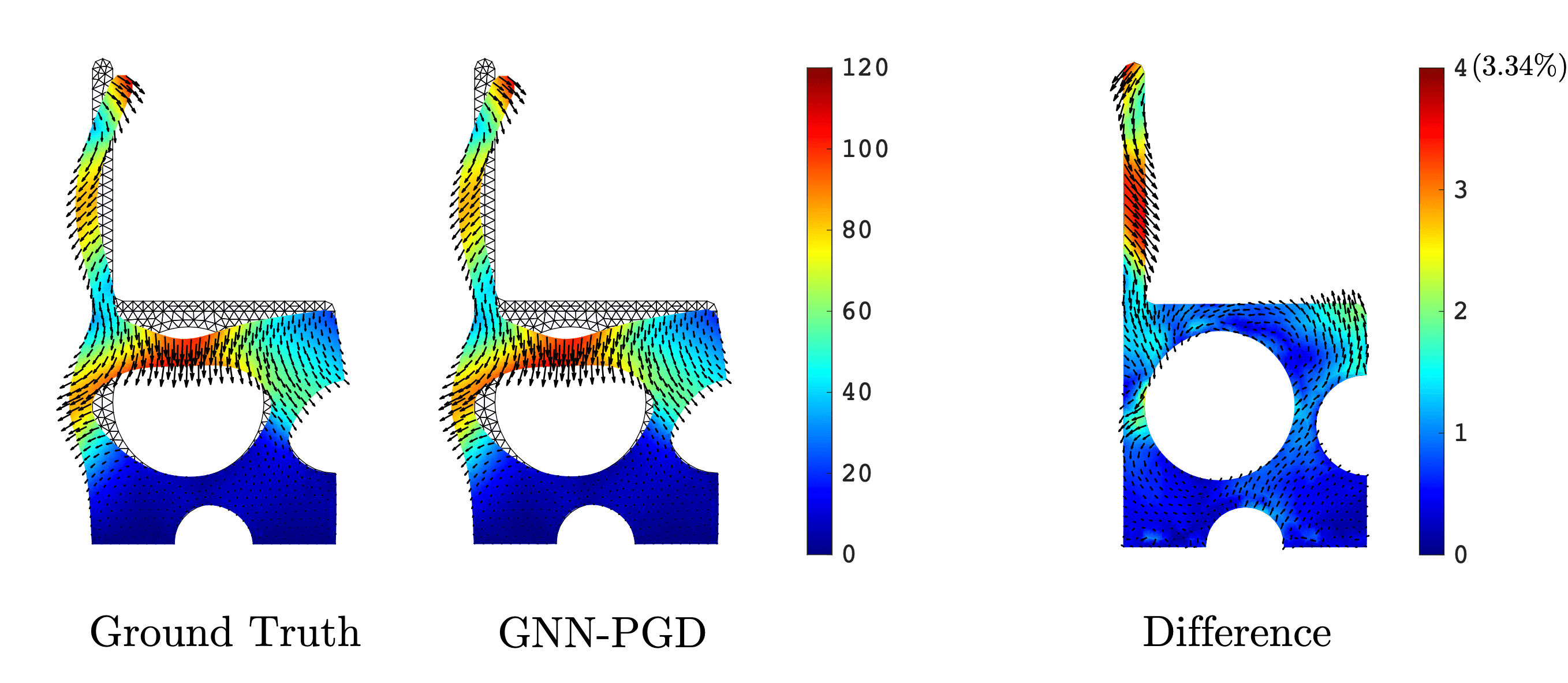}
  \caption{Example of the third mode predicted by GNN-PGD for seat 23 from the test database, together with the ground truth and error.}
  \label{fig:mode3_example}
\end{figure}

\begin{figure}[!ht]
  \center
  \noindent\includegraphics[width=16cm]{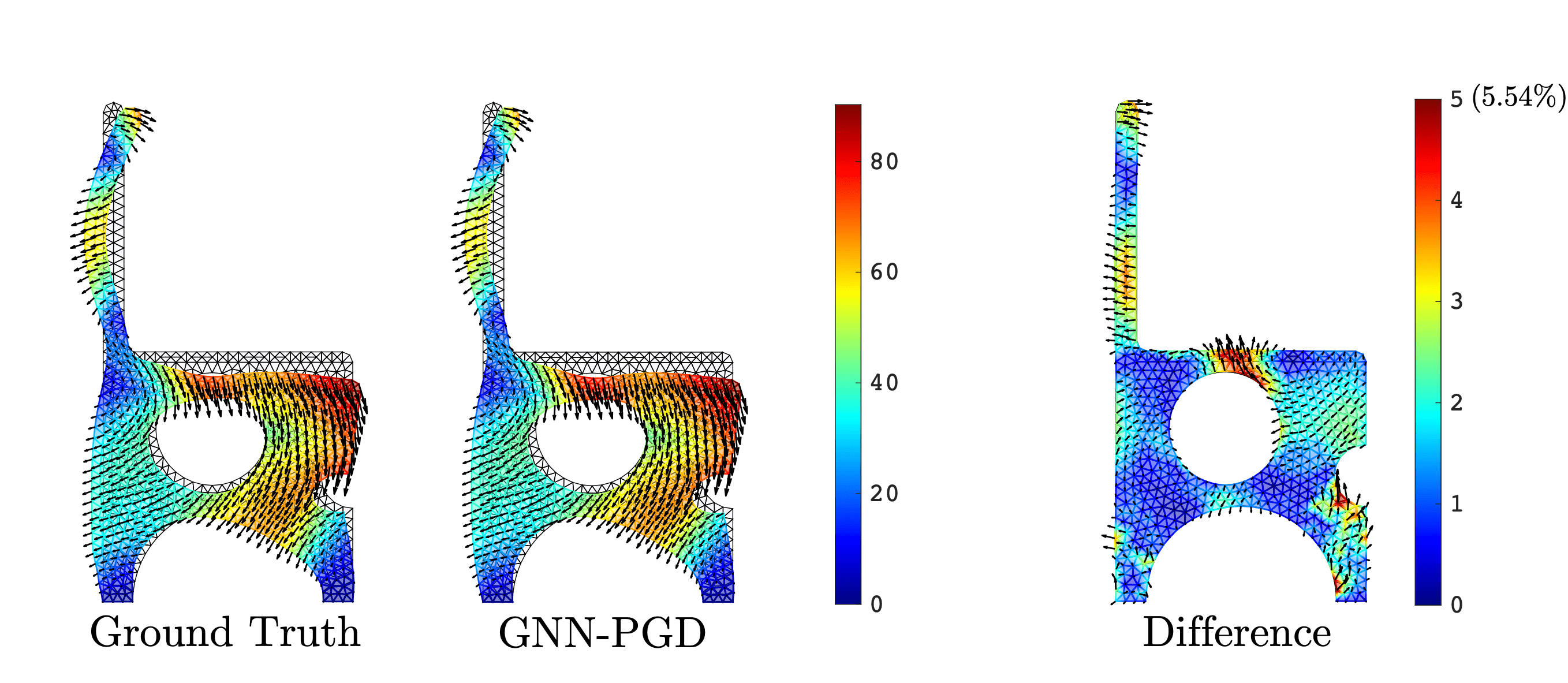}
  \caption{Example of the first mode predicted by GNN-PGD for seat 44 from the test database, together with the ground truth and error.}
  \label{fig:mode33_example}
\end{figure}

\subsubsection{Error of the fully-reconstructed space-time fields} 

Using the ROBs predicted above, we analyze the complete space-time field error associated with the test geometries $\left\{\bm{\Omega}_p^{\text{test}}\right\}_{p=\{1,\dots,50\}}$. As described in \autoref{sec:galerkin}, the temporal component can be obtained from a Galerkin projection of the predicted ROBs $\left\{\vect{P}_p^{\text{GNN}}\right\}_{p=\{1,\dots, 50\}}$. The resulting space-time solutions, denoted  $\left\{\field{U}_p^{\text{GNN}}\right\}_{p=\{1,\dots, 50\}}$, are compared with the FEM reference fields $\left\{\field{U}_p^{\text{test}}\right\}_{p=\{1,\dots, 50\}}$. The resulting normalized space-time $\bm{L}^2 (\bm{\Omega},\mathcal{I})$ errors (i.e., over all space and time) are presented in \autoref{fig:fieldL2}. One can observe $<5\%$ errors for all tested seat geometries $\left\{\bm{\Omega}_p^{\text{test}}\right\}_{p=\{1,\dots,50\}}$. Such performance is promising from an iterative engineering design standpoint, as discussed in the introduction.

\begin{figure}[!ht]
  \center
  \noindent\includegraphics[width=9cm]{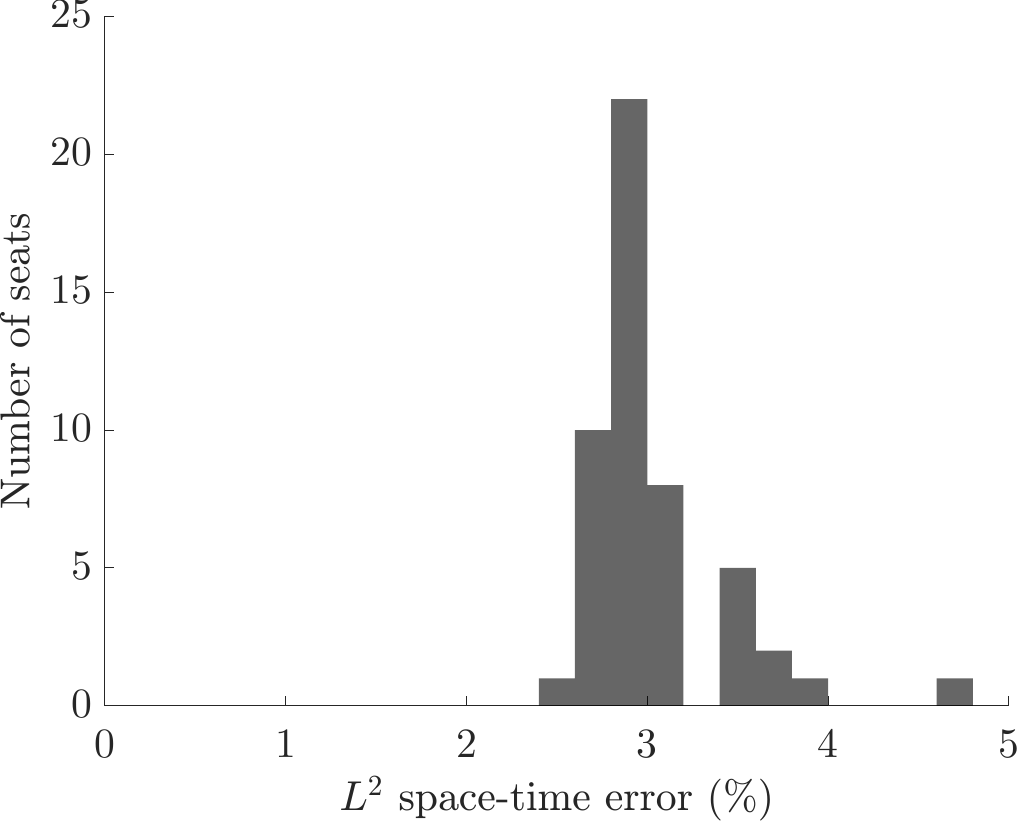}
  \caption{Distribution of the normalized space-time $\bm{L}^2 (\bm{\Omega},\mathcal{I})$ error for the reconstructed fields of the test geometries $\left\{\bm{\Omega}_p^{\text{test}}\right\}_{p=\{1,\dots,50\}}$.}
  \label{fig:fieldL2}
\end{figure}

\subsubsection{Comparison with other approaches}

We compare the results of our proposed GNN-PGD solver with commonly-found autoregressive approaches from literature \cite{hernandez_thermodynamics-informed_2022, pfaff_learning_2021, brandstetter_message_2022}. All other GNN methods, as far as the authors know, construct the temporal solution timestep-by-timestep by feeding back the prediction of the time step $t_i$ as an input to the GNN in order to predict the next time step $t_{i+1}$. A summary of these standard approaches is illustrated in \autoref{fig:autoregressive}.

\begin{figure}[!ht]
  \center
  \noindent\includegraphics[width=14cm]{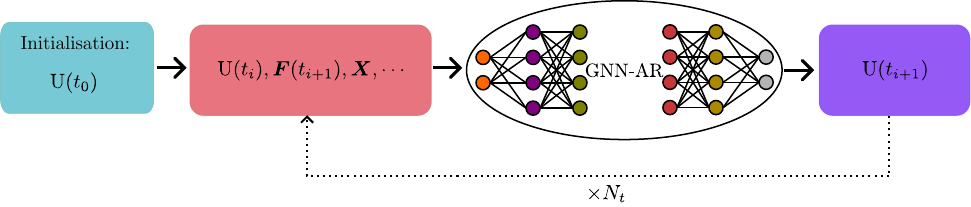}
  \caption{Illustration of a standard GNN-AR autoregressive approach employed as a benchmark comparison for the proposed GNN-PGD method.}
  \label{fig:autoregressive}
\end{figure}

We similarly implemented the standard autoregressive approach (GNN-AR) via pytorch, following \cite{brandstetter_message_2022, pfaff_learning_2021}. The same hyperparameter optimisation strategy (with a DOE, as described in \autoref{sec:gnndetails}) is employed in order to train the GNN-AR for comparative purposes. \autoref{fig:results_compa} presents a comparison of temporal evolution of the spatial $\bm{L}^2$ errors between the GNN-PGD introduced in this work and GNN-AR for all 50 test cases. Results demonstrate a well-known drawback of the timestep-by-timestep GNN-AR approach: an error increase as a function of time (\emph{rollout error increase} \cite{brandstetter_message_2022, pfaff_learning_2021})---for our test cases, up to $15\%$. Conversely, our proposed GNN-PGD approach does not suffer from such an increase in error over time; in fact, it remains bounded with a maximum error of only $3\%$ over all timesteps. Additionally, we find that the learning time (offline time) associated with the GNN-AR approach (16h40) is significantly higher than the GNN-PGD approach (4h25). This is expected since the autoregressive approach considers all $N_t=1000$ snapshots of each geometry in the training database, whereas GNN-PGD only considers the $M=3$ spatial modes associated with these geometries (and reconstructs the temporal component via projection and a computationally-inexpensive resolution of the corresponding ODEs of Equation \eqref{eq:galerkin}). Another advantage of the GNN-PGD approach over the GNN-AR approach is the number of calls to the GNN network. In the case of an autoregressive approach, the number of calls to the network is linked to the number of time steps required in the simulation (e.g., 1000), whereas for the GNN-PGD approach, only one evaluation is required, which translates into a shorter inference time (online time) for the GNN-PGD approach.

\begin{figure}[!ht]
  \center
  \noindent\includegraphics[width=10cm]{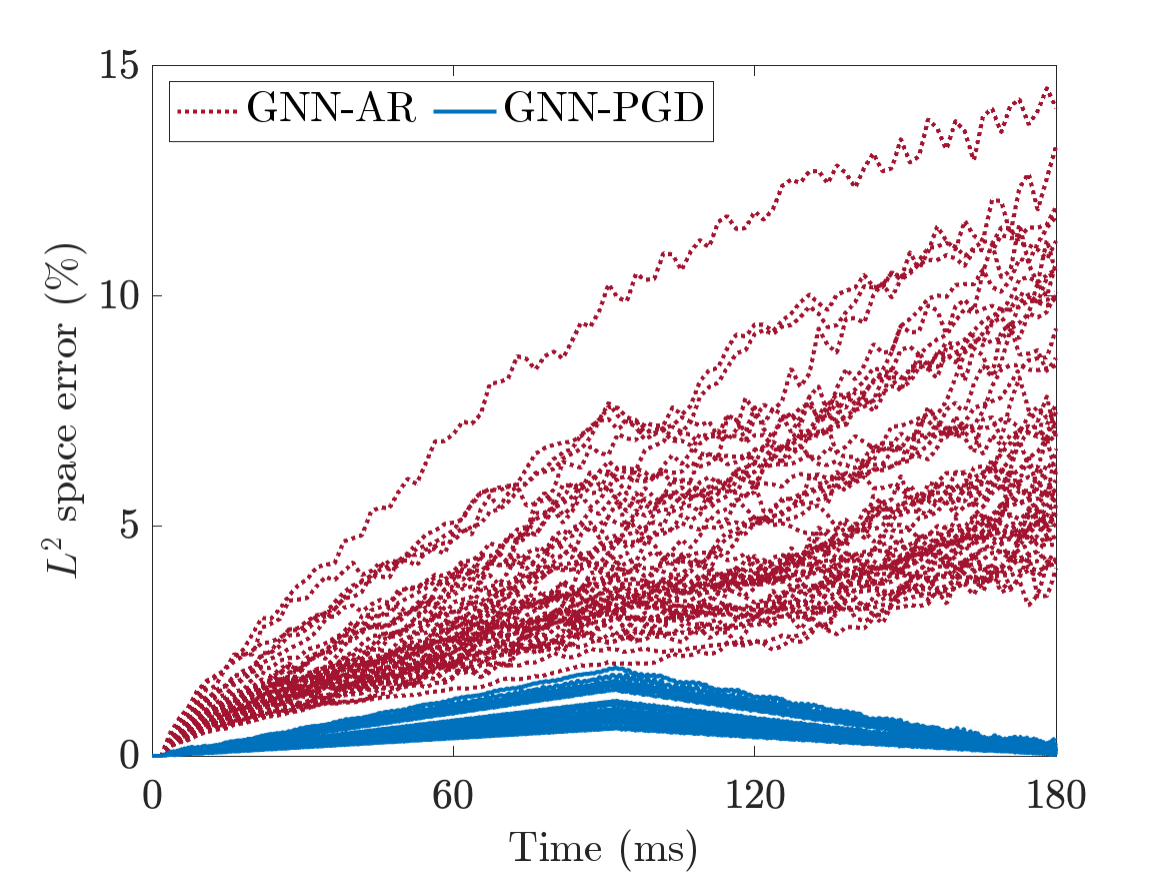}
  \caption{Comparison of the time evolution of the spatial $\bm{L}^2(\bm{\Omega})$ error between the GNN-AR approach~\cite{brandstetter_message_2022, pfaff_learning_2021} and our proposed GNN-PGD approach for the test set $\left\{\bm{\Omega}_p^{\text{test}}\right\}_{p=\{1,\dots,50\}}$.}
  \label{fig:results_compa}
\end{figure}

\subsubsection{Generalization ability of GNN-PGD to infer other topologies}

In order to test GNN-PGD's generalization capabilities for predicting a ROB on topologies that are highly different from those with which it has been trained (including very different discretizations), we consider the seats illustrated in  \autoref{fig:bdd_other}, which contain two or three holes in the interior of the mesh (as opposed to only one used in training, validation, and testing, see \autoref{fig:bdd_gamma}). 
\begin{figure}[!ht]
  \center
  {
  \noindent\includegraphics[height=7cm]{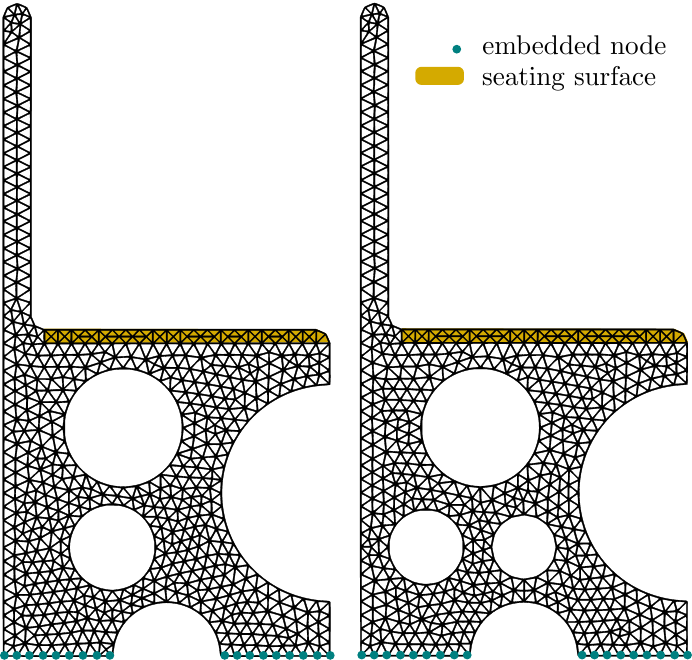}}
  \caption{Example of three seats, exterior to the original database, of very different topologies (i.e., multiple holes).}
\label{fig:bdd_other}
\end{figure}

 Figures \ref{fig:2trous_mode1} illustrates and compares Mode 1 to the ground truth for two circular holes in a seat. Clearly, GNN-PGD is able to capture the general behavior of the mode, with a maximum error of only $15\%$. Similarly, \ref{fig:3trous_mode2} illustrates and compares Mode 2 for a geometry containing three holes, which also indicates a good approximation to the ground truth. A quantification of the $\bm{L}^2$ spatial error for all predicted modes for each of these two cases (two and three holes) is presented in \autoref{tab:otherresults}, along with the complete space-time $\bm{L}^2(\bm{\Omega}, \mathcal{I})$ for the final reconstructed solutions.

\begin{figure}[!ht]
  \center
  \noindent\includegraphics[width=16cm]{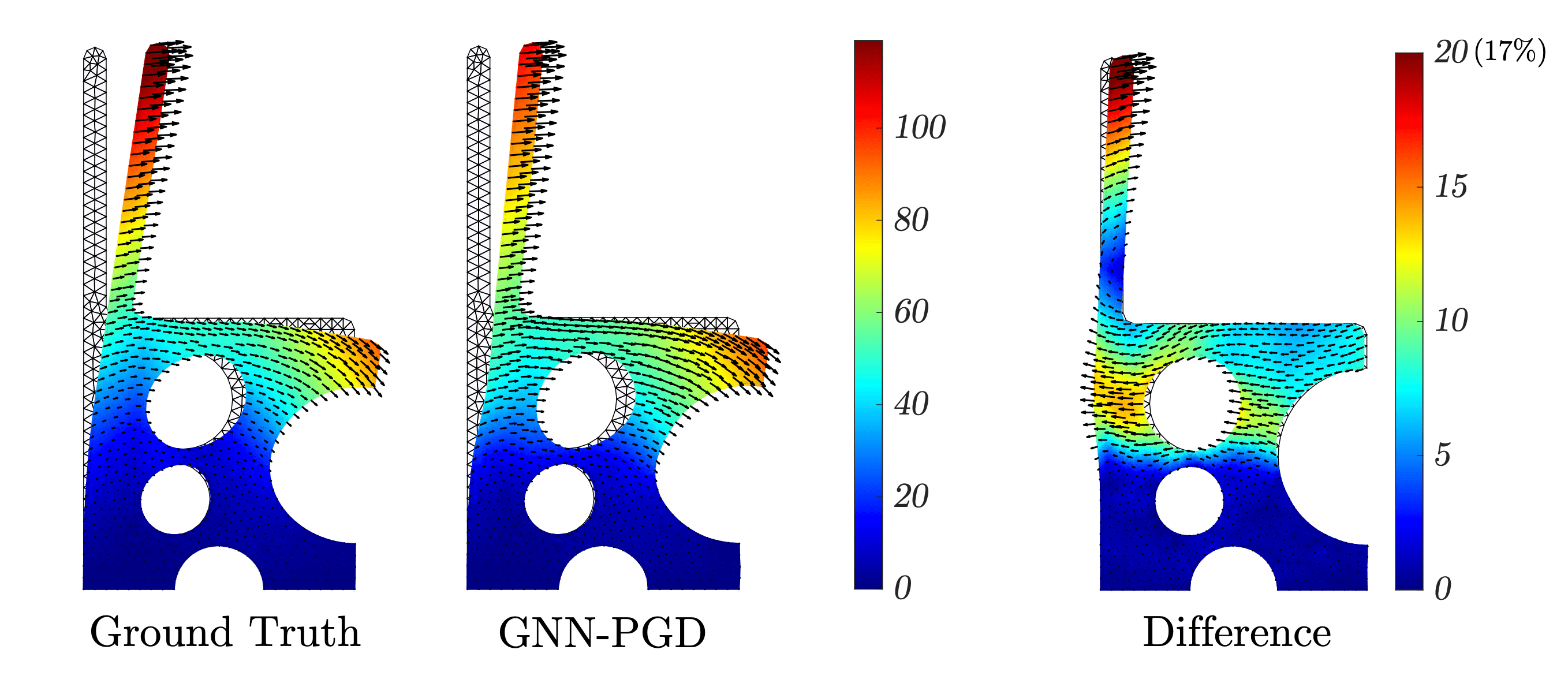}
  \caption{Mode 1 predicted by GNN-PGD (trained on one hole) on a topology with two holes.}
  \label{fig:2trous_mode1}
\end{figure}

\begin{figure}[!ht]
  \center
  \noindent\includegraphics[width=16cm]{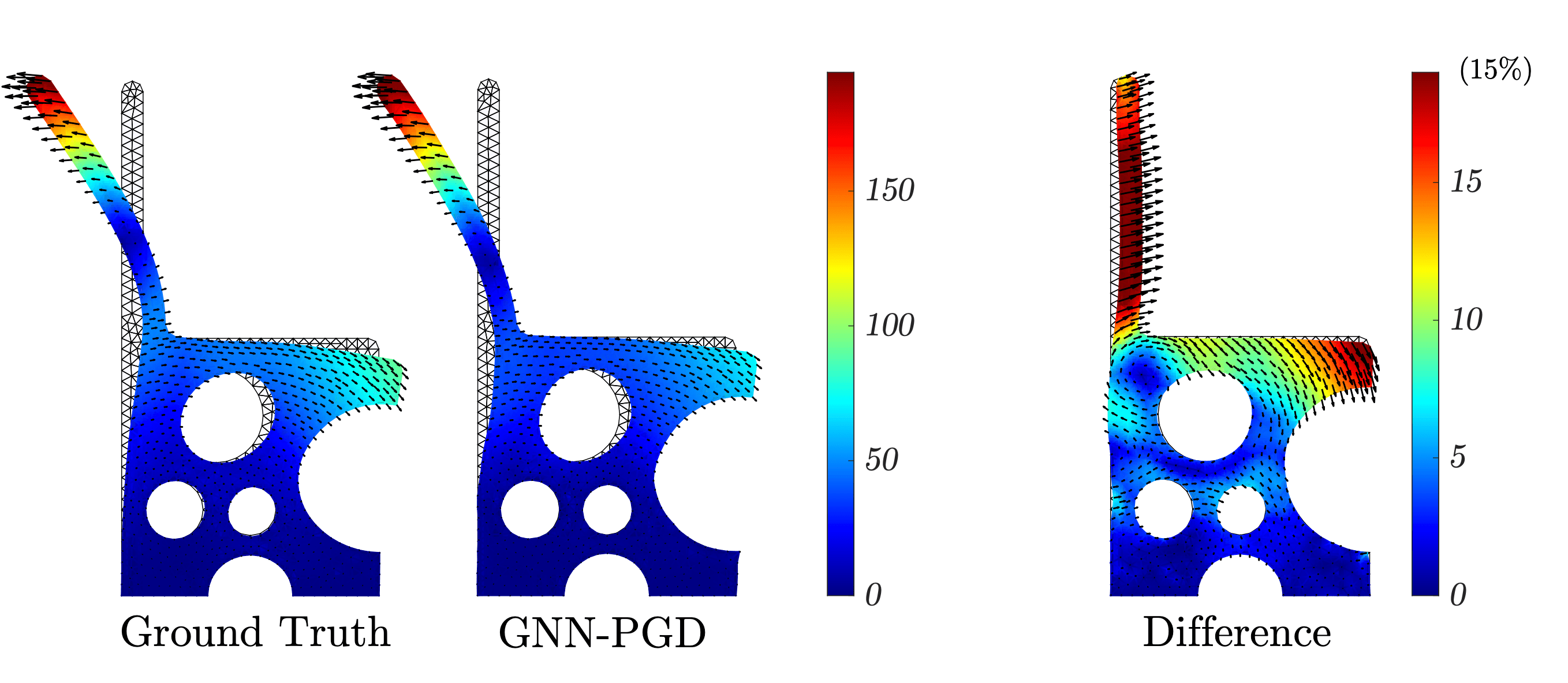}
  \caption{Mode 2 predicted by GNN-PGD (trained on one hole) on a topology with three holes.}
  \label{fig:3trous_mode2}
\end{figure}

\begin{table}[!ht]
\centering
\renewcommand{\arraystretch}{1.5} 
\begin{tabular}{|>{\centering\arraybackslash}m{2cm}|>{\centering\arraybackslash}m{2.5cm}|>{\centering\arraybackslash}m{2.5cm}|>{\centering\arraybackslash}m{2.5cm}|>{\centering\arraybackslash}m{3.5cm}|}
\hline 
\rowcolor{gray!50}
& Mode 1 $\bm{L}^2(\bm{\Omega})$ error  & Mode 2 $\bm{L}^2(\bm{\Omega})$ error   & Mode 3 $\bm{L}^2(\bm{\Omega})$ error & $\mathbf{U}^{\text{GNN}}$ $\bm{L}^2(\bm{\Omega}, \mathcal{I})$ error \\
\hline
\includegraphics[width=0.75cm]{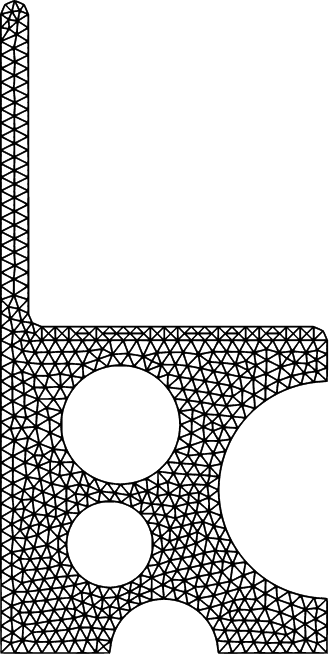} & $16 \%$  & $15 \%$ & $30 \%$ & $16 \%$ \\
\hline
\includegraphics[width=0.75cm]{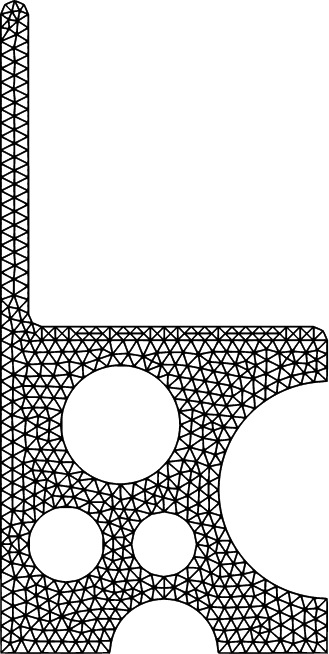}& $17 \%$ & $16 \%$ & $27 \%$ & $18 \%$ \\
\hline
\end{tabular}
\caption{Errors of the GNN-PGD approach for topologies containing more holes than the training data set.}
\label{tab:otherresults}
\end{table}

Although the errors are much higher than those presented for a single hole (on which our GNN-PGD was trained), they are entirely reasonable from an iterative design standpoint. Indeed, for the two new topologies tested, we obtain a final error over the entire space-time field of less than $18 \%$, which demonstrates the remarkable capability of the GNN-PGD solver to generalize to completely different geometries. For pre-dimensioning or pre-sizing of structures, such errors are more than sufficient to approximate various mechanical quantities without having to run full FEM simulations.

\section{Conclusion}

This contribution introduces a new methodology that combines a Galerkin projection-based model reduction method with a deep learning algorithm based on GNNs in order to generate a ROB for an arbitrary geometry of arbitrary discretization size. Results demonstrate that the proposed GNN-PGD solver can produce reduced order models for non-parametric geometries (where usual model reduction tools are limited~\cite{rozza_advances_nodate, courard_pgd-abaques_2016}).  The algorithm proposed and analyzed in this work carries significant advantages over more conventional GNN methods, both in terms of computational cost (training and inference) as well as numerical error.  A case study related to the motivating engineering and industrial applications of this work (structural design) has also been presented, using a reasonably-sized database that results in a GNN-PGD solver that provides low errors that are appropriate for pre-dimensioning or pre-sizing of, e.g., aircraft seats. For such configurations, we have additionally demonstrated a strong capability of the GNN-PGD method to generalize to topologies that are completely different from training and validation databases. 

The proof-of-concept provided in this contribution demonstrates promise for future extensions and other applications. For example, one can incorporate more physics into the GNN-PGD generator so as to be able to generate more modes with even greater accuracy. Towards this, future work is planned for the inclusion of finite element operators, derived from the physical PDEs, as direct inputs to the graph (inspired by \cite{donon_deep_2020}). More importantly, future work entails extending the GNN-PGD method to non-linear problems, e.g., the material non-linearities that are very present in the aeronautical world \cite{tzanakis_structural_2023, cavagna2011neocass, fredriksson1986advanced}. For such contexts, a possible enrichment of the predicted space-time solution (post-processing) may also be required.

\section*{CRediT authorship contribution statement}
\textbf{Victor Matray:} Writing – review \& editing, Writing – original draft,  Validation, Software, Methodology, Investigation, Formal
Analysis, Conceptualization, Visualization. \textbf{Faisal Amlani:} Writing – review \& editing, Writing – original draft, Supervision, Project administration, Conceptualization. \textbf{Frédéric Feyel:} Writing – review \& editing, Supervision, Project administration, Conceptualization. \textbf{David Néron:} Writing – review \& editing, Supervision, Project administration, Conceptualization.
\section*{Acknowledgements}

\noindent  This work was granted access to the HPC resources of IDRIS under the allocation 2023-[AD011014260] made by GENCI. This work was performed using HPC resources from the “Mésocentre” computing center of CentraleSupélec, École Normale Supérieure Paris-Saclay and Université Paris-Saclay supported by CNRS and Région Île-de-France (https://mesocentre.universite-paris-saclay.fr/). Finally, the authors would like to express their sincere gratitude to Safran Tech for supporting this research and for providing the motivating application contexts.

\clearpage

\bibliographystyle{unsrt}  
\bibliography{references}

\end{document}